\def\year{2022}\relax
\documentclass[letterpaper]{article} 
\usepackage{aaai22}  
\usepackage{times}  
\usepackage{helvet}  
\usepackage{courier}  
\usepackage[hyphens]{url}  
\usepackage{graphicx} 
\usepackage{natbib}  
\usepackage{caption}

\usepackage[utf8]{inputenc}
\usepackage{booktabs}
\usepackage{amsfonts}
\usepackage{nicefrac}
\usepackage{microtype}
\usepackage[table]{xcolor}
\usepackage{mathtools,bbm}
\usepackage{color}
\usepackage{multirow}
\usepackage{subcaption}
\usepackage{makecell}
\usepackage{cleveref}
\usepackage{textcomp}
\usepackage{gensymb}
\usepackage{marginnote}
\usepackage{enumitem}

\definecolor{Gray}{gray}{0.9}

\newcommand{\stdv}[1]{\scriptsize$\pm$#1}

\DeclareMathOperator*{\argmax}{arg\,max}

\newcommand{\ie}{\textit{i}.\textit{e}.}
\newcommand{\eg}{\textit{e}.\textit{g}.}

\newcommand{\reviewer}[3]{
	\expandafter\newcommand\csname #1\endcsname[1]{
		\textcolor{#3}{[#2: ##1]}
	}
}

\DeclareCaptionStyle{ruled}{labelfont=normalfont,labelsep=colon,strut=off} 
\usepackage{algorithm}
\usepackage{algorithmic}

\usepackage{newfloat}
\usepackage{listings}
\floatstyle{ruled}
\newfloat{listing}{tb}{lst}{}
\floatname{listing}{Listing}

\title{Consistency Regularization for Adversarial Robustness}
\author{
    Jihoon Tack\textsuperscript{$\dagger$}, Sihyun Yu\textsuperscript{$\dagger$}, Jongheon Jeong\textsuperscript{$\dagger$}, Minseon Kim\textsuperscript{$\dagger$}, Sung Ju Hwang\textsuperscript{$\dagger$,$\ddagger$}, Jinwoo Shin\textsuperscript{$\dagger$}
}
\affiliations{
    \textsuperscript{$\dagger$}Korea Advanced Institute of Science and Technology (KAIST), Daejeon, South Korea\\
    \textsuperscript{$\ddagger$}AITRICS, Seoul, South Korea\\
    \texttt{\{jihoontack,sihyun.yu,jongheonj,minseonkim,sjhwang82,jinwoos\}@kaist.ac.kr}
}

\begin{document}

\maketitle

\begin{abstract}

\emph{Adversarial training} (AT) is currently one of the most successful methods to obtain the adversarial robustness of deep neural networks. However, the phenomenon of robust overfitting, \ie, the robustness starts to decrease significantly during AT, has been problematic, not only making practitioners consider a bag of tricks for a successful training, \eg, early stopping, but also incurring a significant generalization gap in the robustness. In this paper, we propose an effective regularization technique that prevents robust overfitting by optimizing an auxiliary `consistency' regularization loss during AT. Specifically, we discover that data augmentation is a quite effective tool to mitigate the overfitting in AT, and develop a regularization that forces the predictive distributions after attacking from two different augmentations of the same instance to be similar with each other. Our experimental results demonstrate that such a simple regularization technique brings significant improvements in the test robust accuracy of a wide range of AT methods. More remarkably, we also show that our method could significantly help the model to generalize its robustness against unseen adversaries, \eg, other types or larger perturbations compared to those used during training. Code is available at \url{https://github.com/alinlab/consistency-adversarial}.

\end{abstract}

\section{Introduction}

Despite the remarkable success of deep neural networks (DNNs) in real-world applications \cite{he2016deep,girshick2015fast,amodei2016deep}, recent studies have demonstrated that DNNs are vulnerable to adversarial examples, \ie, inputs crafted by an imperceptible perturbation which confuse the network prediction \cite{szegedy2014intriguing, goodfellow2015explaining}. This vulnerability of DNNs raises serious security concerns about their deployment in the real-world applications \cite{kurakin2016adversarial,li2019adversarial}, \eg, self-driving cars and secure authentication system \cite{chen2015deepdriving}.

In this respect, there have been significant efforts to design various defense techniques against the adversarial examples, including input denoising \cite{guo2018countering,liao2018defense}, detection techniques \cite{ma2018characterizing,lee2018simple}, and certificating the robustness of a classifier \cite{cohen2019certified,jeong2020consistency}. Overall, \emph{adversarial training} (AT) is currently one of the most promising ways to obtain the adversarial robustness of DNNs, \ie, directly augmenting the training set with adversarial examples \cite{goodfellow2015explaining,madry2018towards}. Recent studies have been actively investigating a better form of AT \cite{qin2019adversarial, zhang19p, Wang2020Improving}.

\begin{figure}[t]
\centering
\begin{subfigure}{0.225\textwidth}
\centering
\includegraphics[width=\textwidth]{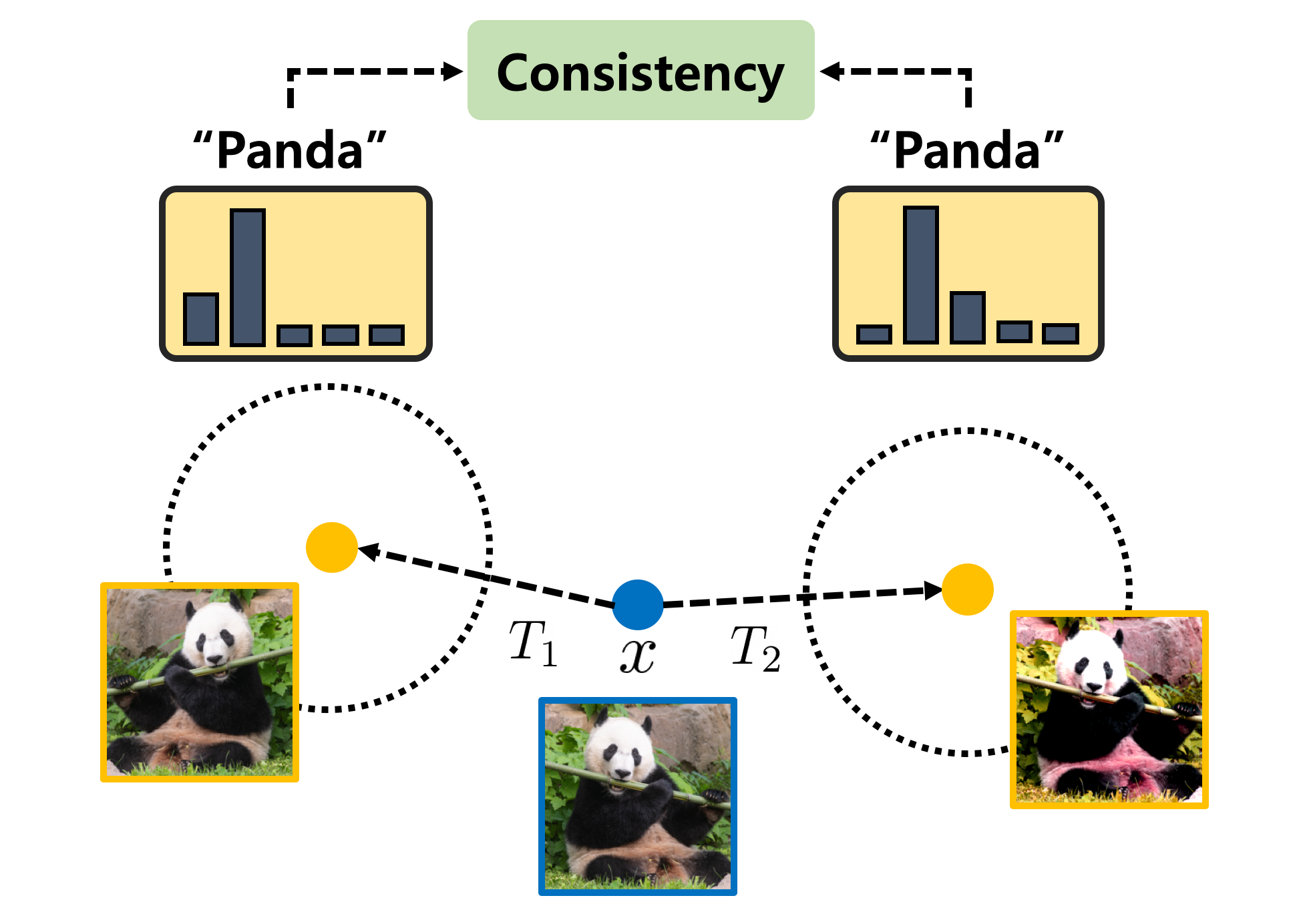}
\caption{Conventional CR}
\end{subfigure}
\begin{subfigure}{0.225\textwidth}
\centering
\includegraphics[width=\textwidth]{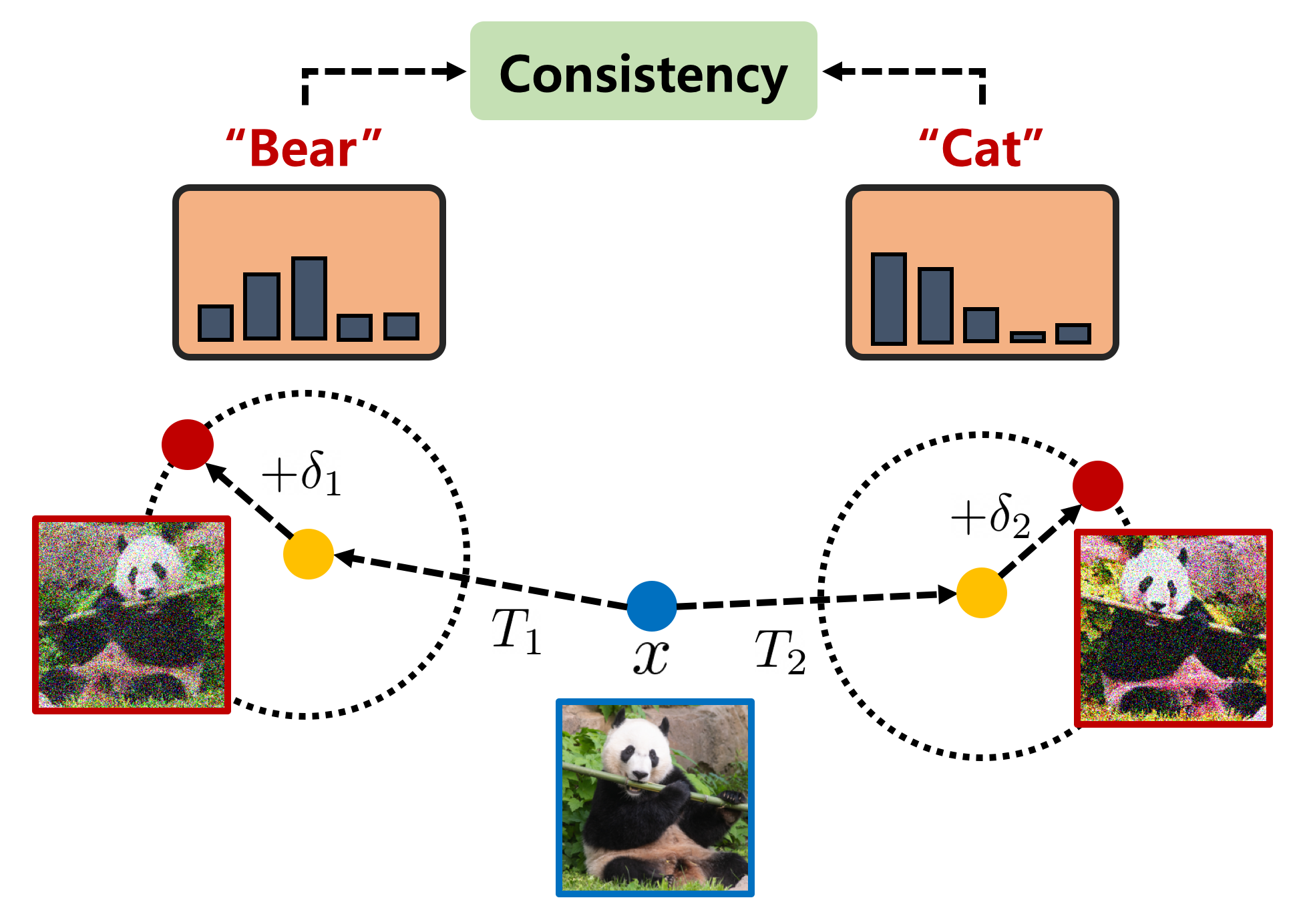}
\caption{Proposed CR}
\end{subfigure}
\caption{
An overview of our consistency regularization (CR) and conventional approach \cite{hendrycks2019augmix,xie2020unsupervised}. Our regularization forces the predictive distribution of \emph{attacked} augmentations to be consistent. $T$ and $\delta$ indicates the randomly sampled augmentation, and the corresponding adversarial noise, respectively.}\label{fig:concept-fig}
\end{figure}

One of the major downsides that most AT methods suffer from, however, is a significant generalization gap of adversarial robustness between the train and test datasets \cite{yang2020adversarial}, possibly due to an increased sample complexity induced by the non-convex, minimax nature of AT \cite{schmidt2018adversarially}. More importantly, it has been observed that such a gap gradually increases from the middle of training \cite{rice2020overfitting}, \ie, overfitting, which makes practitioners to consider several heuristic approaches for a successful optimization \eg, early stopping \cite{zhang19p}. Only recently, a few proposed more advanced regularization techniques, \eg, self-training \cite{chen2021robust}\footnote{We do not consider comparing with the method by \citet{chen2021robust} as they require pre-training additional models.} and weight perturbation \cite{wu2020adversarial}, but it is still largely unknown to the community that why and how only such sophisticated training schemes could be effective to prevent the robust overfitting of AT. 

\textbf{Contribution.} 
In this paper, we suggest to optimize an auxiliary `\emph{consistency}' regularization loss, as a simpler and easy-to-use alternative for regularizing AT. To this end, we first found that the existing data augmentation (DA) schemes are already quite effective to reduce the robust overfitting in AT. Yet, it is contrast to  the recent studies \cite{rice2020overfitting,gowal2020uncovering} which reported DA does not help for AT. Our new finding is that considering more diverse set of augmentations than the current conventional practice can prevent the robust overfitting: we use AutoAugment \cite{cubuk2019autoaugment} which is an effective augmentation for standard cross-entropy training. 

Upon the observation, we claim that the way of incorporating such augmentations could play a significant role in AT. Specifically, we suggest to optimize an auxiliary \emph{consistency regularization} loss during AT: it forces \emph{adversarial examples} from two independent augmentations of the same input to have similar predictions. Here, we remark that forcing the prediction consistency over `clean' DA is widely used for many purposes \cite{zhang2019consistency,hendrycks2019augmix}, however, it looks highly non-trivial at first glance whether matching such attack directions over DA is useful in any sense. Our finding is that the attack direction provides intrinsic information of the sample (other than its label), where the most frequently attacked class is the most confusing class of the `clean' input, \ie, class with the maximum softmax probability disregarding the true label. The proposed regularization loss injects a strong inductive bias to the model that such `dark' knowledge \cite{hinton2015distilling} over DA should be consistent. Our regularization technique is easy to apply to any existing AT methods \cite{madry2018towards,zhang19p,Wang2020Improving}, yet effectively improves the performance.

We verify the efficacy of our scheme through extensive evaluations on CIFAR-10/100 \cite{krizhevsky2009learning} and Tiny-ImageNet.\footnote{\url{https://tiny-imagenet.herokuapp.com/}} Overall, our experimental results show that the proposed regularization can be easily adapted for a wide range of AT methods to prevent overfitting in robustness. For example, our regularization could improve the robust accuracy of WideResNet \cite{zagoruyko2016wide} trained via standard AT \cite{madry2018towards} on CIFAR-10 from 45.62\%$\to$52.36\%. Moreover, we show that our regularization could even notably improve the robustness against unforeseen adversaries \cite{tramer2019adversarial}, \ie, when the adversaries assume different threat models from those used in training: \eg, our method could improve the $l_1$-robustness of TRADES \cite{zhang19p} from 29.58\%$\to$48.32\% on PreAct-ResNet \cite{he2016identity}. Finally, we also observe that our method could be even beneficial for the corruption robustness \cite{hendrycks2019robustness}. 

\section{Consistency Regularization for \\Adversarial Robustness}
\label{sec:method}

In this section, we introduce a simple yet effective strategy for preventing the robust overfitting in adversarial training (AT).
We first review the concept of AT and introduce one of popular AT methods in Section~\ref{sec:at}. We then start in Section~\ref{sec:observation} by showing that the data augmentations can effectively prevent the robustness overfitting. 
Finally, in Section~\ref{sec:consistency-regularization}, we propose a simple yet effective consistency regularization to further utilize the given data augmentations in AT.


\subsection{Preliminaries: Adversarial Training}
\label{sec:at}

We consider a classification task with a given $K$-class dataset $\mathcal{D}=\{(x_i,y_i)\}_{i=1}^{n}\subseteq \mathcal{X} \times \mathcal{Y}$, where $x\in\mathbb{R}^{d}$ represents an input sampled from a certain data-generating distribution $P$ in an \textit{i.i.d.}\ manner, and $\mathcal{Y}:=\{1,\dots,K\}$ represents a set of possible class labels. Let  $f_{\theta}:\mathbb{R}^{d}\to\Delta^{K-1}$ be a neural network modeled to output a probability simplex $\Delta^{K-1}\in\mathbb{R}^{K}$, \eg, via a softmax layer. The notion of adversarial robustness requires $f_{\theta}$ to perform well not only on $P$, but also on the worst-case distribution near $P$ under a certain distance metric. More concretely, the adversarial robustness we primarily focus in this paper is the \emph{$\ell_p$-robustness}: \ie, for a given $p \ge 1$ and a small $\epsilon > 0$, we aim to train a classifier $f_{\theta}$ that correctly classifies $(x+\delta, y)$ for any $\lVert\delta\rVert_{p}\le\epsilon$, where $(x, y)\sim P$.

The high level idea of \emph{adversarial training} (AT) is to directly incorporate adversarial examples to train the classifier \cite{goodfellow2015explaining}, hence the network becomes robust to such adversaries. In general, AT methods formalize the training of $f_\theta$ as an alternative min-max optimization with respect to $\theta$ and $||\delta||_p\le \epsilon$, respectively; \ie, one minimizes a certain classification loss $\mathcal{L}$ with respect to $\theta$ while an adversary maximizes $\mathcal{L}$ by perturbing the given input to $x+\delta$ during training. Here, for a given $\mathcal{L}$, we denote the inner maximization procedure of AT as $\mathcal{L}_{\mathtt{adv}}(x, y; \theta)$: 
\begin{equation}
    \mathcal{L}_{\mathtt{adv}}(x, y; \theta) :=
    \max_{\lVert\delta\rVert_{p}\le\epsilon}
    \mathcal{L}\big(x+ \delta,y ; \theta\big).
    \label{equ:augment-robust-risk}
\end{equation}
For example, one of most basic form of AT method \cite{madry2018towards} considers to design $\mathcal{L}_{\mathtt{adv}}$ with the standard cross-entropy loss $\mathcal{L}_{\mathtt{CE}}$ (we also provide an overview on other types of AT objective such as TRADES \cite{zhang19p} and MART \cite{Wang2020Improving}, in the supplementary material):
\begin{equation}
    \mathcal{L}_{\mathtt{AT}}:=
    \max_{\lVert\delta\rVert_{p}\le\epsilon}
    \mathcal{L}_{\mathtt{CE}}\big(f_{\theta}(x+\delta),y\big).
    \label{equ:madry-at}
\end{equation}

\subsection{Effect of Data Augmentations in Adversarial Training}
\label{sec:observation}

\begin{figure*}[t]
\centering
\begin{subfigure}{0.4\textwidth}
\centering
\includegraphics[width=\textwidth]{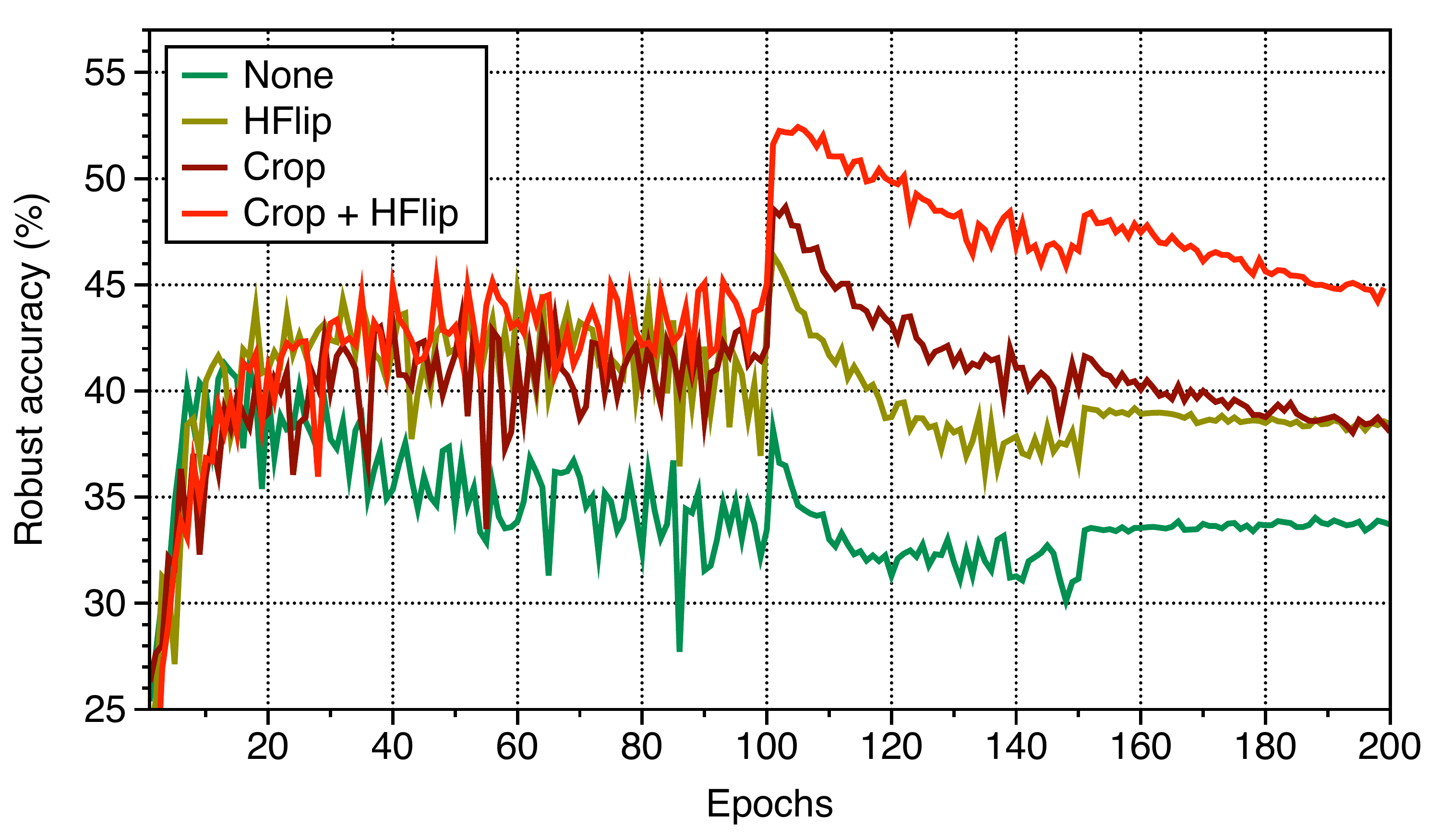}
\caption{Conventional augmentations}
\label{fig:removal-robustness}
\end{subfigure}
~~~~~~~~~~~~
\begin{subfigure}{0.4\textwidth}
\centering
\includegraphics[width=\textwidth]{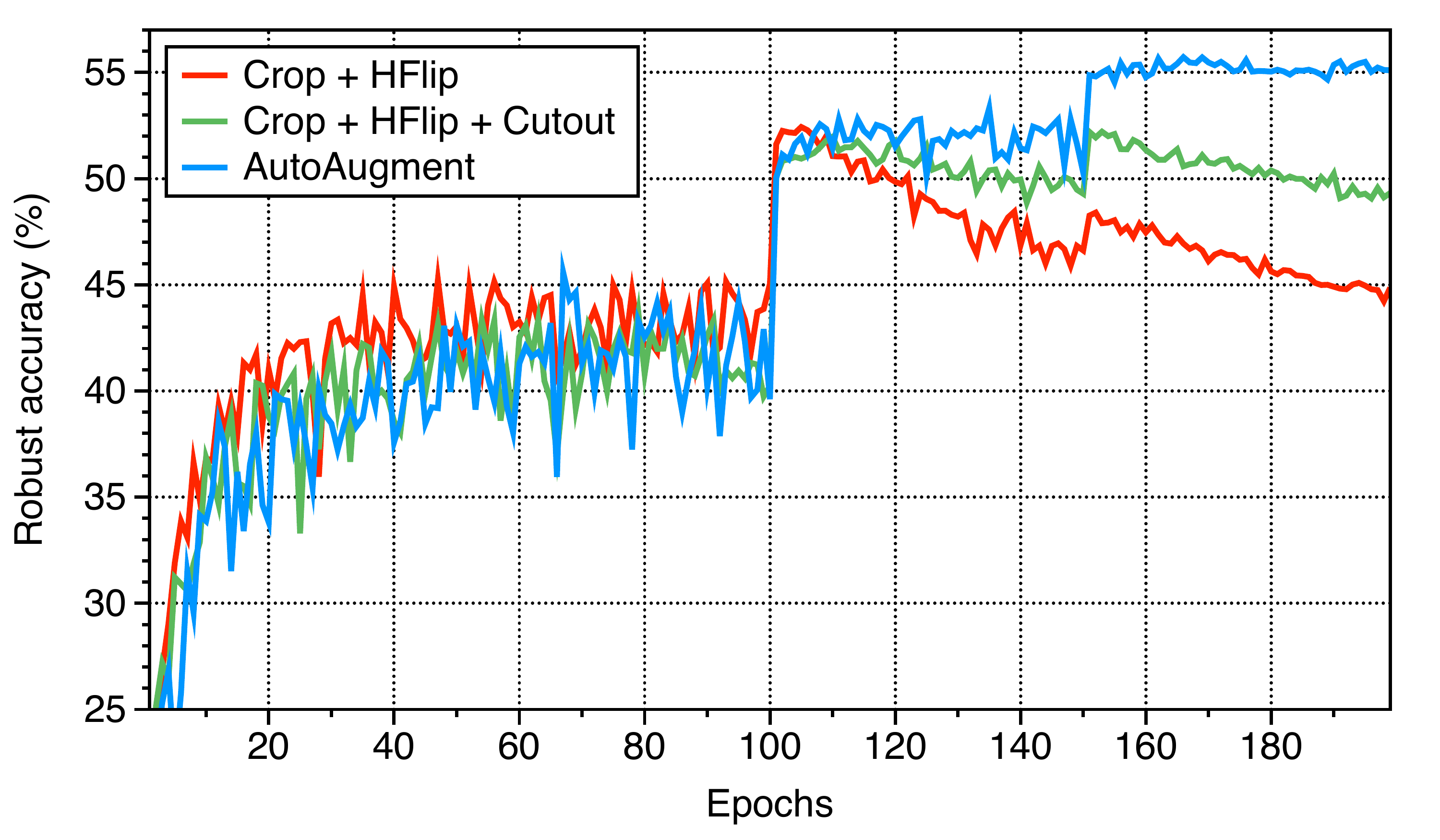}
\caption{Additional augmentations}
\label{fig:addition-robustness}
\end{subfigure}
\caption{
Robust accuracy (\%) against PGD-10 attack on standard AT \cite{madry2018towards} under (a) conventional augmentations, and (b) additional augmentations to the convention. We consider PreAct-ResNet-18 trained on CIFAR-10. We use $l_{\infty}$ threat model with $\epsilon=8/255$. None, HFlip, and Crop, indicates no augmentation, horizontal flip, and random crop, respectively. Note that the AutoAugment \cite{cubuk2019autoaugment} includes horizontal flip, random crop and Cutout \cite{devries2017improved}. The jump in robust accuracy at 100, 150 epochs is due to a drop in the learning rate.
}
\label{fig:observation}
\end{figure*}

Now, we investigate the utility of data augmentations in AT. We first show that current standard choices of augmentation in AT are already somewhat useful for relaxing the robust overfitting, where considering more diverse augmentations is even more effective. Throughout this section, we train PreAct-ResNet-18 \cite{he2016identity} on CIFAR-10 \cite{krizhevsky2009learning} using standard AT \cite{madry2018towards}, following the training details of \citet{rice2020overfitting}. We use projected gradient descent (PGD) with 10 iterations under $\epsilon=8/255$ (step size of $2/255$) with $l_{\infty}$ constraint to perform adversarial attacks for both training and evaluation. Formally, for a given training sample $(x,y)\sim\mathcal{D}$, and augmentation $T\sim\mathcal{T}$, the training loss is:
\begin{equation}
    \max_{||\delta||_{\infty} \leq \epsilon} \mathcal{L}_{\mathtt{CE}} (f_{\theta}(T(x) + \delta), y).
    \label{equ:madry-at-aug}
\end{equation}
Unless otherwise specified, we assume the set of baseline augmentations $\mathcal{T}:=\mathcal{T}_{\mathtt{base}}$ (\ie, random crop with 4 pixels zero padding and horizontal flip) by default for this section.

%

\textbf{Role of base augmentations in adversarial training.}
We recognize the set of base augmentations $\mathcal{T}_{\mathtt{base}}$ has been commonly used in most existing AT methods, and observe these augmentations are
already somewhat useful for relaxing the robust overfitting in AT.
To this end, we conduct a controlled experiment by removing each augmentation from the pre-defined augmentation set $\mathcal{T}_{\mathtt{base}}$ and train the network. 
%
%
Figure \ref{fig:removal-robustness} summarizes the result of the experiment. As each augmentation is removed, not only the robustness degrades, but also the adversarial overfitting is getting significant. This phenomenon stands out more when no augmentations are applied during AT, which only shows the increment of robust accuracy at the first 5\% of the whole training procedure. 
This result implies that there may exist an augmentation family that effectively prevents the robust overfitting as the base augmentation is already useful.  

\textbf{Reducing robust overfitting with data augmentations.} 
We further find that the existing data augmentation schemes
are already quite effective to reduce the robust overfitting in AT.
Specifically, we utilize AutoAugment \cite{cubuk2019autoaugment} which is the state-of-the-arts augmentation scheme for the standard cross-entropy training.
As shown in Figure \ref{fig:addition-robustness}, the robust overfitting is gradually reduced as more diverse augmentations are used, and even the best accuracy improves. Note that AutoAugment is more diverse than the conventional augmentations as it includes the $\mathcal{T}_{\mathtt{base}}$ and Cutout \cite{devries2017improved}. Interestingly, our empirical finding somewhat shows a different conclusion from the previous studies \cite{gowal2020uncovering} which conclude that data augmentations are not effective for preventing the robust overfitting. We further discuss a detailed analysis of data augmentations in the supplementary material.


\subsection{Consistency Regularization for Adversarial Training}
\label{sec:consistency-regularization}

We suggest to optimize a simple auxiliary \emph{consistency regularization} during AT to further utilize the given data augmentations. Specifically, our regularization forces \emph{adversarial examples} from two independent augmentations of an instance to have a similar prediction (see Figure \ref{fig:concept-fig}). 
However, it is highly non-trivial whether matching such attack directions via consistency regularization is useful, which we essentially investigate in this paper.
Our major finding is that the attack direction itself contains intrinsic information of the instance, as in Section \ref{sec: exp-ablation}. For example, the most frequently attacked class is the most confusing class of the `clean' input, \ie, class with the maximum softmax probability disregarding the true label. Hence, our regularization utilize this dark knowledge (other than the true labels) of samples and induce a strong inductive bias to the classifier.

Formally, for a given data point $(x,y)\sim\mathcal{D}$ and augmentations $T_{1},T_{2}\sim\mathcal{T}$, we denote $\delta_{i}$ as an adversarial noise of $T_{i}(x)$, \ie, $\delta_{i}:=\argmax_{\lVert\delta\rVert_{p}\le\epsilon}\mathcal{L}\big(T_i(x),y,\delta ; \theta\big)$. We consider regularizing the temperature-scaled distribution $\hat{f}_{\theta}(x;\tau)$ \cite{guo2017calibration} over the adversarial examples across augmentations to be consistent, where $\tau$ is the temperature hyperparameter. Concretely, temperature-scaled classifier is $\hat{f}_\theta(x;\tau)=\mathtt{Softmax}(z_{\theta}(x)/\tau)$ where $z_{\theta}(x)$ is the logit value of $f_{\theta}(x)$, \ie, activation before the softmax layer of $f_{\theta}(x)$. Then the proposed regularization loss is given by:
\begin{equation}
    \mathtt{JS}\Big(
    \hat{f}_{\theta}\big(T_{1}(x)+\delta_{1};\tau\big) \parallel \hat{f}_{\theta}\big(T_{2}(x)+\delta_{2};\tau\big)
    \Big),
    \label{equ:consistency}
\end{equation}
where $\mathtt{JS}(\cdot\parallel\cdot)$ denotes the Jensen-Shannon divergence. Since the augmentations are randomly sampled in every training step, adversarial example's predictions become consistent regardless of augmentation selection when minimizing the proposed objective. We note that the motivation behind the temperature scaling is that the confidence of prediction (\ie, maximum softmax value) is relatively low on AT than the standard training. Hence, we compensate this issue by enforcing the sharp distribution by using a small temperature.

\begin{table}[ht]
\caption{
Comparison of the consistency regularization (CR) loss. We report clean accuracy and robust accuracy (\%) against PGD-100 attack of PreAct-ResNet-18 trained on CIFAR-10. We use $l_{\infty}$ threat model with $\epsilon=8/255$.
}\centering\small
\begin{tabular}{lcc}
\toprule
Loss & Clean & PGD-100 \\
\midrule
AT \eqref{equ:madry-at-aug} & 85.41 & 55.18 \\
AT \eqref{equ:madry-at-aug} + previous CR \eqref{equ:conven-consistency} & 88.01 & 53.11 \\
AT \eqref{equ:madry-at-aug} + proposed CR \eqref{equ:consistency} & 86.45 & 56.38 \\
\bottomrule
\end{tabular}
\label{tab:type-consistency}
\end{table}

\textbf{Comparison to other consistency regularization loss over DA.}
There has been prior works that suggested a consistency regularization loss to better utilize DA \cite{hendrycks2019augmix,zhang2019consistency,sohn2020fixmatch}, 
which can be expressed with the following form:
\begin{equation}
    \mathtt{D}\Big(
    f_{\theta}\big(T_{1}(x)\big),f_{\theta}\big(T_{2}(x)\big)
    \Big),
    \label{equ:conven-consistency}
\end{equation}
where $\mathtt{D}$ is a discrepancy function.
The regularization term used in \eqref{equ:conven-consistency} 
has a seemingly similar formula to ours but there is a fundamental difference: our method \eqref{equ:consistency}
does not match the predictions directly for the `clean' augmented samples, but does after \emph{attacking} them independently, \ie, $f_{\theta}(T(x)+\delta)$.
To examine which one is better, we compare \eqref{equ:consistency} with \eqref{equ:conven-consistency} under the same discrepancy function, $\mathtt{D}:=\mathtt{JS}$ and same augmentation family, \ie, AutoAugment. As shown in Table \ref{tab:type-consistency}, our design choice \eqref{equ:consistency} improves both clean and robust accuracy compare to the baseline \eqref{equ:madry-at-aug}, while the prior consistency regularization \eqref{equ:conven-consistency} shows significant degradation on the robust accuracy. We additionally try to attack only single augmented instance in \eqref{equ:conven-consistency}, where it also shows degradation in the robust accuracy, \eg, 53.20\% against PGD-100 (such regularization is used in unsupervised AT \cite{kim2020adversarial}).

\textbf{Overall training objective.} 
In the end, we derive a final training objective, $\mathcal{L}_{\mathtt{total}}$: an AT objective combined with the consistency regularization loss \eqref{equ:consistency}. To do so, we consider the average of inner maximization objective on AT $\mathcal{L}_{\mathtt{adv}}$ \eqref{equ:augment-robust-risk} over two independent augmentations $T_1, T_2 \sim\mathcal{T}$, as minimizing (\ref{equ:augment-robust-risk}) over the augmentations $T\sim\mathcal{T}$ is equivalent to an average of (\ref{equ:augment-robust-risk}) over $T_1$ and $T_2$:
\begin{equation}
\frac{1}{2}\Big(\mathcal{L}_{\mathtt{adv}} \big(T_1(x), y; \theta\big) 
               +\mathcal{L}_{\mathtt{adv}} \big(T_2(x), y; \theta\big)\Big).
\label{eq:mean-obj}
\end{equation}
We then combine our regularizer \eqref{equ:consistency} with a given hyperparameter $\lambda$, into the average of inner maximization objectives \eqref{eq:mean-obj}. Then the final training objective $\mathcal{L}_{\mathtt{total}}$ is as follows:
\begin{equation*}
\begin{split}
\mathcal{L}_{\mathtt{total}} &:= 
\frac{1}{2}\sum_{i=1}^{2}\mathcal{L}_{\mathtt{adv}} \big(T_{i}(x), y; \theta\big) \\
 &+\lambda\cdot\mathtt{JS}\Big(
    \hat{f}_{\theta}\big(T_{1}(x)+\delta_{1};\tau\big) \parallel \hat{f}_{\theta}\big(T_{2}(x)+\delta_{2};\tau\big)
    \Big).
\end{split}
\end{equation*}
Note that our regularization scheme is agnostic to the choice of AT objective, hence, can be easily incorporated into well-known AT methods \cite{madry2018towards,zhang19p,Wang2020Improving}. For example, considering standard AT loss \cite{madry2018towards} as the AT objective, \ie, $\mathcal{L}_{\mathtt{adv}}=\mathcal{L}_{\mathtt{AT}}$ \eqref{equ:madry-at}, the final objective becomes:
\begin{equation*}
\begin{split}
\mathcal{L}_{\mathtt{total}} &= 
\frac{1}{2} \sum_{i=1}^{2} 
\max_{\lVert\delta_i\rVert_{p}\le\epsilon}\mathcal{L}_{\mathtt{CE}}\big(f_{\theta}(T_{i}(x)+\delta_i),y\big)\\
&+ 
\lambda\cdot\mathtt{JS}\Big(
\hat{f}_{\theta}\big(T_{1}(x)+\delta_{1};\tau\big) \parallel 
\hat{f}_{\theta}\big(T_{2}(x)+\delta_{2};\tau\big)
\Big).
\label{equ:objective-all-at}
\end{split}
\end{equation*}

We introduce explicit forms of other variants of final objective $\mathcal{L}_{\mathtt{total}}$ for other AT methods, \eg, TRADES \cite{zhang19p} and MART \cite{Wang2020Improving}, integrated with our regularization loss, in the supplementary material.

\section{Experiments} \label{sec: exp}

We verify the effectiveness of our technique on image classification datasets: CIFAR-10/100 \cite{krizhevsky2009learning} and Tiny-ImageNet. Our results exhibit that incorporating simple consistency regularization scheme into the existing adversarial training (AT) methods significantly improve adversarial robustness against various attacks \cite{carlini2017towards, madry2018towards, croce2020reliable}, including data corruption \cite{hendrycks2019robustness}. Intriguingly, our method shows better robustness against \textit{unseen} adversaries compared to other baselines. Moreover, our method somewhat surpass the performance of the recent regularization technique \cite{wu2020adversarial}. Finally, we perform an ablation study to validate each component of our approach.


\subsection{Experimental Setups}
\textbf{Training details.}
We use PreAct-ResNet-18 \citep{he2016identity} architecture in all experiments, and additionally use WideResNet-34-10 \cite{zagoruyko2016wide} for white-box adversarial defense on CIFAR-10. For the data augmentation, we consider AutoAugment \cite{cubuk2019autoaugment} where random crop (with 4 pixels zero padding), random horizontal flip (with 50\% of probability), and Cutout \cite{devries2017improved} (with half of the input width) are included. We set the regularization parameter $\lambda=1.0$ in all cases except for applying on WideResNet-34-10 with TRADES and MART where we use $\lambda=2.0$. The temperature is fixed to $\tau=0.5$ in all experiments.

For other training setups, we mainly follow the hyperparameters suggested by the previous studies \cite{pang2021bag,rice2020overfitting}. In detail, we train the network for 200 epochs\footnote{Our method maintains almost the same robust accuracy under the same computational budget to the baselines: reduce the training steps in half. See the supplementary material for more discussion.} using stochastic gradient descent with momentum 0.9, and weight decay of 0.0005. The learning rate starts at 0.1 and is dropped by a factor of 10 at 50\%, and 75\% of the training progress. For the inner maximization for all AT, we set the $\epsilon=8/255$, step size $2/255$, and 10 number of steps with $l_{\infty}$ constraint (see the supplementary material for the $l_{2}$ constraint AT results).

Throughout the section, we mainly report the results where the clean accuracy converges, \ie, {fully trained model}, to focus on the robust overfitting problem \cite{rice2020overfitting}. Nevertheless, we also note that our regularization method achieves better best robust accuracy compare to the AT methods (see Table \ref{tab:1-main-white-all}). 


\begin{table*}[t]
\caption{Clean accuracy and robust accuracy (\%) against white-box attacks of networks trained on various image classification benchmark datasets. All threat models are $l_{\infty}$ with $\epsilon=8/255$. Values in parenthesis denote the result of the checkpoint with the best PGD-10 accuracy, where each checkpoint is saved per epoch. We compare with the baselines trained under random crop and flip. The bold indicates the improved results by our proposed loss.}
\resizebox{\textwidth}{!}{
\small
\centering
\begin{tabular}{c l | ccccc}
    \toprule
    \makecell{Dataset\\(Architecture)} & Method & Clean & PGD-20 & PGD-100 & CW$_{\infty}$ & AutoAttack \\
    \midrule
    \multirow{7}{*}{\makecell{CIFAR-10\\(PreAct-ResNet-18)}}  
    & Standard \cite{madry2018towards} 
    & 84.57 \footnotesize{(83.43)} & 45.04 \footnotesize{(52.82)} 
    & 44.86 \footnotesize{(52.67)} & 44.31 \footnotesize{(50.66)} 
    & 40.43 \footnotesize{(47.63)} \\
    & \textbf{+ Consistency} 
    & \textbf{86.45} \footnotesize{(85.25)} 
    & \textbf{56.51} \footnotesize{(57.53)} 
    & \textbf{56.38} \footnotesize{(57.39)} 
    & \textbf{52.45} \footnotesize{(52.70)} 
    & \textbf{48.57} \footnotesize{(49.05)} \\
    \cmidrule{2-7}
    & {TRADES} \cite{zhang19p} 
    & 82.87 \footnotesize{(82.13)} & 50.95 \footnotesize{(53.98)} 
    & 50.83 \footnotesize{(53.85)} & 49.30 \footnotesize{(51.71)} 
    & 46.32 \footnotesize{(49.32)} \\
    & \textbf{+ Consistency}     
    & \textbf{83.63} \footnotesize{(83.55)} 
    & \textbf{55.00} \footnotesize{(55.16)} 
    & \textbf{54.89} \footnotesize{(54.98)} 
    & \textbf{49.91} \footnotesize{(50.67)} 
    & \textbf{47.68} \footnotesize{(49.01)} \\
    \cmidrule{2-7}
    & {MART} \cite{Wang2020Improving} 
    & 82.63 \footnotesize{(77.00)} 
    & 51.12 \footnotesize{(54.83)} 
    & 50.91 \footnotesize{(54.74)} 
    & 46.92 \footnotesize{(49.26)} 
    & 43.46 \footnotesize{(46.74)} \\
    & \textbf{+ Consistency} 
    & \textbf{83.43} \footnotesize{(81.89)} 
    & \textbf{59.59} \footnotesize{(60.48)} 
    & \textbf{59.52} \footnotesize{(60.47)} 
    & \textbf{51.78} \footnotesize{(51.83)} 
    & \textbf{48.91} \footnotesize{(48.95)} \\
    \midrule
    \multirow{7}{*}{\makecell{CIFAR-10\\(WideResNet-34-10)}}  
    & Standard \cite{madry2018towards} 
    & 86.37 \footnotesize{(87.55)} & 50.16 \footnotesize{(55.86)}
    & 49.80 \footnotesize{(55.65)} & 49.25 \footnotesize{(54.45)}
    & 45.62 \footnotesize{(51.24)} \\
    & \textbf{+ Consistency} 
    & \textbf{89.82} \footnotesize{(89.93)} 
    & \textbf{58.63} \footnotesize{(61.11)} 
    & \textbf{58.41} \footnotesize{(60.99)} 
    & \textbf{56.38} \footnotesize{(57.80)} 
    & \textbf{52.36} \footnotesize{(54.08)} \\
    \cmidrule{2-7}
    & TRADES \cite{zhang19p} 
    & 85.05 \footnotesize{(84.30)} & 51.20 \footnotesize{(57.34)}
    & 50.89 \footnotesize{(57.20)} & 50.88 \footnotesize{(55.08)}
    & 46.17 \footnotesize{(53.02)} \\
    & \textbf{+ Consistency}
    & \textbf{87.71} \footnotesize{(87.92)} 
    & \textbf{58.39} \footnotesize{(59.12)} 
    & \textbf{58.19} \footnotesize{(58.99)} 
    & \textbf{54.84} \footnotesize{(55.97)} 
    & \textbf{51.94} \footnotesize{(53.11)} \\
    \cmidrule{2-7}
    & {MART} \cite{Wang2020Improving} 
    & 85.75 \footnotesize{(83.98)} & 49.31 \footnotesize{(57.28)}
    & 49.06 \footnotesize{(57.22)} & 48.05 \footnotesize{(53.21)}
    & 44.96 \footnotesize{(50.62)} \\
    & \textbf{ + Consistency} 
    & \textbf{87.17} \footnotesize{(85.81)} 
    & \textbf{63.26} \footnotesize{(64.95)} 
    & \textbf{62.81} \footnotesize{(64.80)} 
    & \textbf{57.46} \footnotesize{(56.24)} 
    & \textbf{52.41} \footnotesize{(53.33)} \\
    \midrule
    \multirow{2}{*}{\makecell{CIFAR-100\\(PreAct-ResNet-18)}}
    & Standard \cite{madry2018towards}  
    & 57.13 \footnotesize{(57.10)} & 22.36 \footnotesize{(29.67)} 
    & 22.25 \footnotesize{(29.65)} & 21.97 \footnotesize{(27.99)} 
    & 19.85 \footnotesize{(25.38)} \\
    & \textbf{+ Consistency} 
    & \textbf{62.73} \footnotesize{(61.62)} 
    & \textbf{30.75} \footnotesize{(32.33)} 
    & \textbf{30.62} \footnotesize{(32.24)} 
    & \textbf{27.63} \footnotesize{(28.39)} 
    & \textbf{24.55} \footnotesize{(25.52)} \\
    \midrule
    \multirow{2}{*}{\makecell{Tiny-ImageNet\\(PreAct-ResNet-18)}} 
    & Standard \cite{madry2018towards} 
    & 41.54 \footnotesize{(45.26)} & 11.71 \footnotesize{(20.92)} 
    & 11.60 \footnotesize{(20.87)} & 11.20 \footnotesize{(18.72)} 
    & 9.63 \footnotesize{(16.03)} \\
    & \textbf{+ Consistency}     
    & \textbf{50.15} \footnotesize{(49.46)} 
    & \textbf{21.33} \footnotesize{(23.31)} 
    & \textbf{21.24} \footnotesize{(23.24)} 
    & \textbf{19.08} \footnotesize{(20.29)} 
    & \textbf{15.69} \footnotesize{(16.90)} \\
    \bottomrule
\end{tabular}
}
\label{tab:1-main-white-all}
\end{table*}

\subsection{Main Results}
\textbf{White-box attack.} 
We consider a wide range of white-box adversarial attacks, in order to extensively measure the robustness of trained models without gradient obfuscation \cite{athalye2018obfuscated}: PGD \cite{madry2018towards} with 20 and 100 iterations (step size with $2\epsilon/k$, where $k$ is the iteration number), CW$_{\infty}$ \cite{carlini2017towards}, and AutoAttack \cite{croce2020reliable}.\footnote{We regard AutoAttack as a white-box attack, while it both includes white-box and black-box attacks. See the supplementary material for black-box transfer attack results. We use the official code for the AutoAttack: \url{https://github.com/fra31/auto-attack}.} We report the fully trained model's accuracy and the result of the checkpoint with the best PGD accuracy (of 10 iterations), where each checkpoint is saved per epoch.

As shown in Table \ref{tab:1-main-white-all}, incorporating our regularization scheme into existing AT methods consistently improves both best and last white-box accuracies against various adversaries across different models and datasets. The results also demonstrates that our method effectively prevents robust overfitting as the gap between the best and last accuracies has been significantly reduced in all cases. In particular, for TRADES with WideResNet-34-10, our method's robust accuracy gap under AutoAttack is only 1.17\%, while the baseline's gap is 6.85\%, which is relatively 6 times smaller. More intriguingly, consideration of our regularization technique into the AT methods boosts the clean accuracy as well in all cases. We notice that such improvement is non-trivial, as some works have reported a trade-off between a clean and robust accuracies in AT \cite{tsipras2018robustness, zhang19p}. 

\begin{table*}[t]
\caption{Robust accuracy (\%) of PreAct-ResNet-18 trained with $l_{\infty}$ of $\epsilon=8/255$ constraint against unseen attacks. For unseen attacks, we use PGD-100 under different sized $l_{\infty}$ balls, and other types of norm ball, \eg, $l_{1},l_{2}$. We compare with the baselines trained under random crop and flip. The bold indicates the improved results by the proposed method.}
\begin{center}
\small
\begin{tabular}{c l cc cc cc}
        \toprule
        & & \multicolumn{2}{c}{$l_{\infty}$} & \multicolumn{2}{c}{$l_{2}$} & \multicolumn{2}{c}{$l_{1}$} \\
        \cmidrule(r){3-4} \cmidrule(r){5-6} \cmidrule(r){7-8}
        Dataset & Method\ $\backslash\ \epsilon$ & $4/255$ & $16/255$ & $150/255$ & $300/255$ & $2000/255$ & $4000/255$ \\
        \midrule
        \multirow{7}{*}{CIFAR-10}  
        & Standard \cite{madry2018towards}
        & 65.93 & 19.23 & 52.56 & 25.68 & 45.96 & 36.85 \\
        & \textbf{+ Consistency}
        & \textbf{73.74} & \textbf{23.47} & \textbf{65.81} & \textbf{36.87} & \textbf{58.66} & \textbf{50.79} \\
        \cmidrule{2-8}
        & TRADES \cite{zhang19p} 
        & 68.30 & 24.17 & 56.14 & 28.94 & 44.08 & 29.58 \\
        & \textbf{+ Consistency} 
        & \textbf{70.33} & \textbf{26.52} & \textbf{63.70} & \textbf{39.16} & \textbf{56.48} & \textbf{48.32} \\
        \cmidrule{2-8}
        & MART \cite{Wang2020Improving}
        & 67.76 & 23.36 & 57.17 & 30.98 & 46.61 & 34.63 \\
        & \textbf{+ Consistency}
        & \textbf{72.67} & \textbf{30.31} & \textbf{66.17} & \textbf{43.76} & \textbf{60.57} & \textbf{54.19} \\
        \midrule
        \multirow{2}{*}{CIFAR-100}  
        & Standard \cite{madry2018towards}
        &  36.14 & 7.37 & 27.97 & 11.98 & 30.48 & 27.29 \\
        & \textbf{+ Consistency}
        & \textbf{46.11} & \textbf{11.53} & \textbf{39.77} & \textbf{20.69} & \textbf{36.04} & \textbf{32.75} \\
        \midrule
        \multirow{2}{*}{Tiny-ImageNet}  
        & Standard \cite{madry2018towards}
        & 23.23 & 2.69 & 28.05 & 17.80 & 33.30 & 31.55 \\
        & \textbf{+ Consistency}
        & \textbf{34.18} & \textbf{5.74} & \textbf{40.06} & \textbf{30.62} & \textbf{43.90} & \textbf{42.65} \\
        \bottomrule
\end{tabular}
\label{tab:2-main-unseen}
\end{center}
\end{table*}

\begin{table}[!t]
\caption{Mean corruption error (mCE) (\%) of PreAct-ResNet-18 trained on CIFAR-10, and tested with CIFAR-10-C dataset \cite{hendrycks2019robustness}. The arrow on the right side of the evaluation metric indicates the descending order of the value is better. We compare with the baselines trained under random crop and flip. The bold indicates the improved results by the proposed method.}
\centering
\small
\begin{tabular}{l c}
    \toprule
    Method  & mCE $\downarrow$  \\
    \midrule
    Standard cross-entropy & 27.02 \\
    \midrule
    Standard \cite{madry2018towards} &  24.03 \\
    \textbf{+ Consistency} & \textbf{21.83} \\
    \midrule
    TRADES \cite{zhang19p} & 25.50 \\
    \textbf{+ Consistency} & \textbf{23.95} \\
    \midrule
    MART \cite{Wang2020Improving} & 26.20 \\
    \textbf{+ Consistency} & \textbf{24.41} \\
    \bottomrule
\end{tabular}
\label{tab:4-main-mCE}
\end{table}

\textbf{Unseen adversaries.} 
We also evaluate our method against \emph{unforeseen} adversaries, \eg, robustness on different attack radii of $\epsilon$, or even on different norm constraints of $l_2$ and $l_1$, as reported in Table \ref{tab:2-main-unseen}. We observe that combining our regularization method could consistently and significantly improve the robustness against all the considered unseen adversaries tested. It is remarkable that our method is especially effective against $l_1$ adversaries compared to the baselines, regarding the fundamental difficulty of achieving the mutual robustness against both $l_1$ and $l_{\infty}$ attacks \cite{tramer2019adversarial, Croce2020Provable}. Hence, we believe our regularization scheme can also be adapted to AT methods for training robust classifiers against multiple perturbations \cite{tramer2019adversarial,maini2020adversarial}.

\textbf{Common corruption.}
We also validate the effectiveness of our method on corrupted CIFAR-10 dataset \cite{hendrycks2019robustness}, \ie, consist of 19 types of corruption such as snow, zoom blur. We report the mean corruption error (mCE) of each model in Table \ref{tab:4-main-mCE}. The results show that the mCE consistently improves combined with our regularization loss regardless of AT methods. Interestingly, our method even reduces the error (from the standard cross-entropy training) of corruptions that are not related to the applied augmentation or noise, \eg, zoom blur error 25.8\%$\to$19.8\%. We note that common corruption is also important and practical defense scenario \cite{hendrycks2019robustness}, therefore, obtaining such robustness should be a desirable property for a robust classifier.


\begin{table*}[!t]
\caption{
Clean accuracy and robust accuracy (\%) against diverse attacks of each individual, and combined regularization. The numbers below the attack methods, indicate the radius of the perturbation $\epsilon$. All results are reported on PreAct-ResNet-18 trained under various image classification benchmark datasets. The bold indicates the best results.
}
\resizebox{\textwidth}{!}{
\centering
\begin{tabular}{c l c ccc cc cc}
    \toprule
    & & & \multicolumn{3}{c}{$l_{\infty}$ (Seen)} & \multicolumn{2}{c}{$l_{2}$ (Unseen)} &  \multicolumn{2}{c}{$l_{1}$ (Unseen)} \\
    \cmidrule(r){4-6} \cmidrule(r){7-8} \cmidrule(r){9-10}
    Dataset & Method & Clean & \makecell{PGD-100\\\small{(8/255)}} &
    \makecell{CW$_{\infty}$\\\small{(8/255)}} &
    \makecell{AutoAttack\\\small{(8/255)}} & \makecell{PGD-100\\\small{(150/255)}} &  \makecell{PGD-100\\\small{(300/255)}} &  \makecell{PGD-100\\\small{(2000/255)}} &  \makecell{PGD-100\\\small{(4000/255)}} \\
    \midrule
    \multirow{3}{*}{CIFAR-10}  
    & Standard \cite{madry2018towards} & 84.57 & 44.86 & 44.31 & 40.43 & 52.56 & 25.68 & 45.96 & 36.85 \\
    & {+ AWP} \cite{wu2020adversarial} & 80.34 & 55.39 & 52.31 & \textbf{49.60} & 61.39 & 36.05 & 56.30 & 48.37 \\
    & \textbf{+ Consistency} & \textbf{86.45} & \textbf{56.38} & \textbf{52.45} & {48.57} & \textbf{65.81} & \textbf{36.87} & \textbf{58.66} & \textbf{50.79} \\
    \midrule
    \multirow{3}{*}{CIFAR-100}
    & Standard \cite{madry2018towards}  & 56.96 & 20.86 & 21.20 & 18.93 & 27.65 & 11.08 & 26.49 & 21.48 \\
    & {+ AWP} \cite{wu2020adversarial} & 52.91 & 30.06 & 26.42 & 24.32 & 35.71 & 20.18 & 33.63 & 30.38 \\
    &  \textbf{+ Consistency} & \textbf{62.73} & \textbf{30.62} & \textbf{27.63} & \textbf{24.55} & \textbf{39.77} & \textbf{20.69} & \textbf{36.04} & \textbf{32.75} \\
    \midrule
    \multirow{3}{*}{Tiny-ImageNet}
    & Standard \cite{madry2018towards}  & 41.54 & 11.60 & 11.20 & 9.63 & 28.05 & 17.80 & 33.30 & 31.55 \\
    & {+ AWP} \cite{wu2020adversarial} & 40.25 & 20.64 & 18.05 & 15.26 & 33.31 & 26.86 & 35.48 & 34.22 \\
    &  \textbf{+ Consistency} & \textbf{50.15} & \textbf{21.24} & \textbf{19.08} & \textbf{15.69} & \textbf{40.06} & \textbf{30.62} & \textbf{43.90} & \textbf{42.65} \\
    \bottomrule
\end{tabular}
\label{tab:5-main-compare}
}\end{table*}

\subsection{Comparison with \citet{wu2020adversarial}} 
\label{sec:recent-regul}

In this section, we consider a comparison with Adversarial weight perturbation (AWP) \cite{wu2020adversarial}\footnote{We use the official code: \url{https://github.com/csdongxian/AWP}}, another recent work which also addresses the overfitting problem of AT by regularizing the flatness of the loss landscape with respect to weights via an adversarial perturbation on both input and weights. We present two experimental scenarios showing that our method can work better than AWP. 

\begin{figure}[t]
\centering
\includegraphics[width=0.37\textwidth]{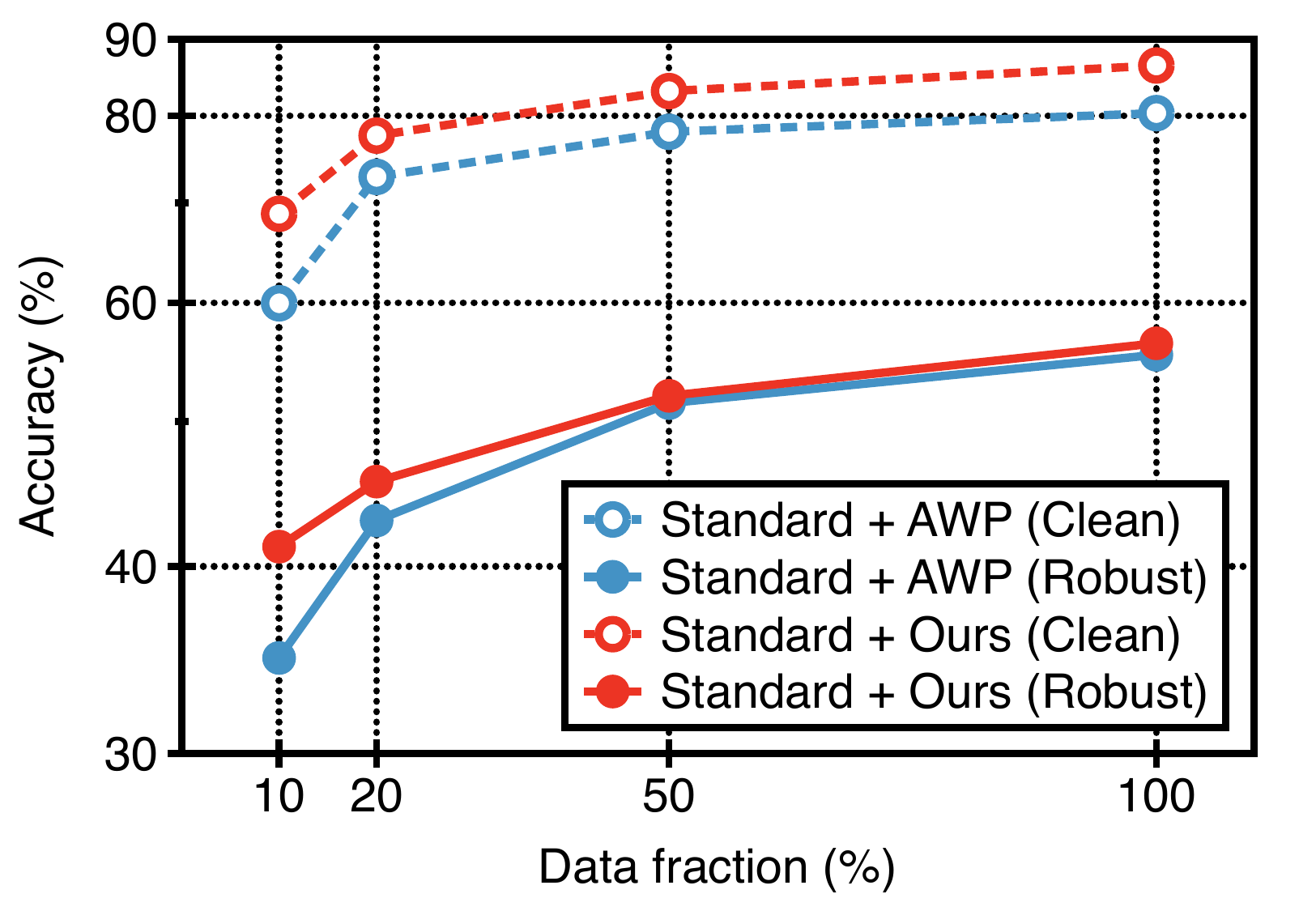}
\caption{ 
Clean accuracy and robust accuracy (\%) against PGD-100 attack of $l_{\infty}$ with $\epsilon=8/255$, under different fraction (\%) of CIFAR-10. We train PreAct-ResNet-18 with AWP \cite{wu2020adversarial} and consistency regularization loss based on standard AT \cite{madry2018towards}.
}
\label{fig:limited-data}
\end{figure}

\textbf{White-box attack and unseen adversaries.} We consider various white-box attacks and unseen adversaries for measuring the robustness. As shown in Table \ref{tab:5-main-compare}, our method shows better results than AWP in $l_{\infty}$ defense in most cases, and outperforms in all cases of unseen adversaries defense, \eg, $l_{2},l_{1}$ constraint attack. In particular, our regularization technique consistently surpass AWP in the defense against the $l_{1}$ constraint attack. In addition, our method shows consistent improvement in clean accuracy, while AWP somewhat suffers from the trade-off between clean and robust accuracy. 

\textbf{Training with limited data.} We also demonstrate that our method is data-efficient: when only a small number of training points are accessible for training the classifier. To this end, we reduce the training dataset's fraction to 10\%, 20\%, and 50\% and train the classifier in each situation. As shown in Figure \ref{fig:limited-data}, our method shows better results compare to AWP, especially learning from the small sized dataset, as our method efficiently incorporates the rice space of data augmentations. In particular, our method obtained 41.2\% robust accuracy even in the case when only 10\% of the total dataset is accessible (where AWP achieves 34.7\%). We note such efficiency is worthy for practitioners, since in such cases, validation dataset for early stopping is insufficient.


\subsection{Ablation Study} 
\label{sec: exp-ablation}
We perform an ablation study on each of the components in our method. Throughout the section, we apply our method to the standard AT \cite{madry2018towards} and use PGD with 100 iterations for the evaluation. We also provide more analysis on the temperature hyperparameter and design choice of consistency regularization in the supplementary material.

\begin{table}[!t]
\caption{Ablation study on each component of our proposed training objective. Reported values are the robust accuracy (\%) against PGD-100 attack of $l_{\infty}$ with $\epsilon=8/255$, 
and mean corruption error (mCE) (\%) of PreAct-ResNet-18 under CIFAR-10. The bold indicates the best result.}
\small\centering
\begin{tabular}{l cc}
    \toprule
    Method  & PGD-100 & mCE $\downarrow$ \\
    \midrule
    Standard \cite{madry2018towards} & 44.86 & 24.03 \\
    + Cutout \cite{devries2017improved} & 49.95 & 24.05 \\
    + AutoAugment \cite{cubuk2019autoaugment} & 55.18 & 23.38 \\
    \textbf{+ Consistency} & \textbf{56.38} & \textbf{22.06} \\
    \bottomrule
\end{tabular}
\label{tab:5-abla-component}
\end{table}

\begin{figure*}[t]

\centering
\includegraphics[width=0.9\textwidth]{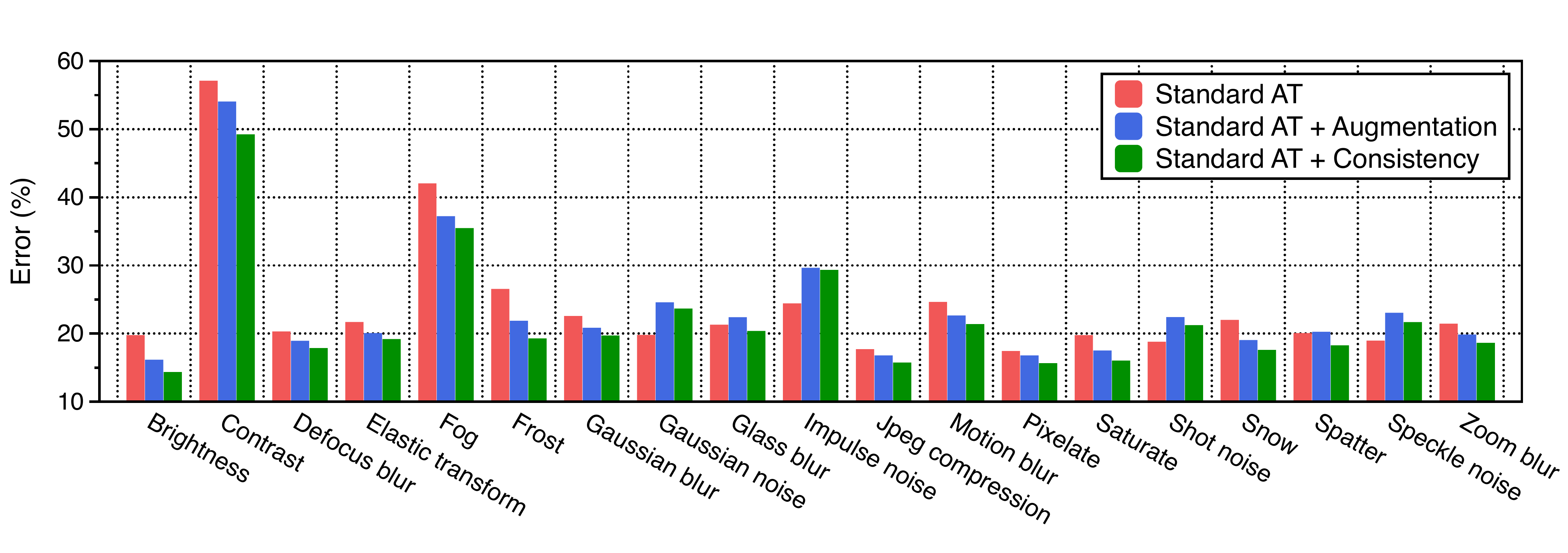}

\caption{
Classification error (\%) on each corruption type of CIFAR-10-C \cite{hendrycks2019robustness} where the $x$-axis labels denote the corruption type. Reported values are measured on PreAct-ResNet-18 trained under standard AT \cite{madry2018towards}, standard AT with AutoAugment \cite{cubuk2019autoaugment}, standard AT with consistency regularization, respectively.
}
\label{fig:mce-bar-plot}
\end{figure*}

\textbf{Component analysis.} 
We perform an analysis on each component of our method, namely the use of (a) data augmentations, and (b) the consistency regularization loss, by comparing their robust accuracy and mean corruption error (mCE). The results in Table \ref{tab:5-abla-component} demonstrate each component is indeed effective, as the performance improves step by step with the addition of the component. We note that the proposed regularization method could not only improve the robust accuracy but also significantly improve the mCE. As shown in Figure \ref{fig:mce-bar-plot}, simply applying augmentation to the standard AT can reduce the error in many cases (13 types out of 19 corruptions) and even reduce the error of corruptions that are not related to the applied augmentation (\eg, motion blur, zoom blur). More interestingly, further adapting the consistency regularization loss can reduce the corruption error in all cases from the standard AT with augmentation. It suggests that the consistency prior is indeed a desirable property for classifiers to obtain robustness (for both adversarial and corruption).

\textbf{Analysis on attack directions.}
To analyze the effect of our regularization scheme, we observe the attacked directions of the adversarial examples. We find that the most confusing class of the `clean' input, is highly like to be attacked. Formally, we define the most confusing class of the given sample $(x,y)$ as $\argmax_{k\neq y}f_{\theta}^{(k)}(x)$ where $f_{\theta}^{(k)}$ is the softmax probability of class $k$. We observe that 77.45\% out of the misclassified adversarial examples predicts the most confusing class. This result implies that the attack direction itself contains the dark knowledge of the given input \citep{hinton2015distilling}, which supports our intuition to match the attack direction.

\section{Conclusion} \label{sec: discussion}

In this paper, we propose a simple yet effective regularization technique to tackle the robust overfitting in adversarial training (AT). Our regularization forces the predictive distributions after attacking from two different augmentations of the same input to be similar to each other. Our experimental results demonstrate that the proposed regularization brings significant improvement in various defense scenarios including unseen adversaries.

\section*{Acknowledgements}
We thank Jaeho Lee, Sangwoo Mo, and Soojung Yang for providing helpful feedbacks and suggestions in preparing an earlier version of the manuscript. This work was supported by Institute of Information \& communications Technology Planning \& Evaluation (IITP) grant funded by the Korea government(MSIT) (No.2019-0-00075, Artificial Intelligence Graduate School Program(KAIST)) and the Engineering Research Center Program through the National Research Foundation of Korea (NRF) funded by the Korean Government MSIT (NRF-2018R1A5A1059921).

\bibliography{reference}


\appendix
\onecolumn
\clearpage
\begin{center}{\bf {\LARGE Supplementary Material:}}
\end{center}
\begin{center}{\bf {\Large Consistency Regularization for Adversarial Robustness}}
\end{center}
\vspace{0.1in}

\section{Overview on Other Adversarial Training Objectives} \label{sec-supple:objective}

In this section, we provide an overview of other types of adversarial training (AT) objectives: TRADES \cite{zhang19p}, and MART \cite{Wang2020Improving}. We use the same notation as in Section \ref{sec:method}.

\subsection{TRADES} 
\citet{zhang19p} showed that there can exist a trade-off between clean and adversarial accuracy and decomposed the objective of standard AT into clean and robust objectives. 
By combining two objectives with a balancing hyperparameter,
one can control such trade-offs.
For a given sample $(x,y)\sim\mathcal{D}$, and a classifier $f_{\theta}$ the proposed training objective is:
\begin{equation}
    \mathcal{L}_{\mathtt{TRADES}}:=
    \mathcal{L}_{\mathtt{CE}}\big(f_{\theta}(x),y\big)
    +    
    \beta \cdot \max_{\lVert\delta\rVert_{p}\le\epsilon}\mathtt{KL}\big(f_{\theta}(x) \parallel f_{\theta}(x+\delta)\big),
    \label{equ:trades}
\end{equation}
where $\mathtt{KL}(\cdot\parallel\cdot)$ denotes the Kullback–Leibler divergence, $\mathcal{L}_{\mathtt{CE}}$ is the cross-entropy loss, and $\beta$ is the hyperparameter to control the trade-off between clean accuracy and robust accuracy. For all experiments, we set the hyperparameter $\beta=6$ by following the original paper.

\noindent\textbf{TRADES with consistency regularization.} By utilizing $\mathcal{L}_{\mathtt{TRADES}}$ as the base AT objective, one can also adapt our regularization scheme for the final objective $\mathcal{L}_{\mathtt{total}}^{\mathtt{TRADES}}$. We also apply the consistency regularization loss between adversarial examples with the Jensen-Shannon divergence, $\mathtt{JS}(\cdot\parallel\cdot)$. For given sampled data augmentations $T_{1},T_{2}\sim\mathcal{T}$ and sharpening temperature $\tau$, the final objective is as follows:
\begin{equation}
\begin{split}
\mathcal{L}_{\mathtt{total}}^{\mathtt{TRADES}} = 
\frac{1}{2} \sum_{i=1}^{2}
\mathcal{L}_{\mathtt{CE}}\big(f_{\theta}(T_{i}(x)),y\big) 
&+    
\beta \cdot \mathtt{KL}\big(f_{\theta}(T_{i}(x)) \parallel f_{\theta}(T_{i}(x)+\delta_{i})\big)
\\
& + \lambda \cdot 
\mathtt{JS} \Big(
\hat{f}_{\theta}\big(T_{1}(x)+\delta_{1};\tau\big) \parallel 
\hat{f}_{\theta}\big(T_{2}(x)+\delta_{2};\tau\big)
\Big)
,
\label{equ:objective-all-trades}
\end{split}
\end{equation}
where $\delta_{i} = \argmax_{\lVert\delta\rVert_{p}\le\epsilon}\mathtt{KL}\big(f_{\theta}(T_{i}(x)) \parallel f_{\theta}(T_{i}(x)+\delta)\big)$, and $\hat{f}_{\theta}(x;\tau)$ is the temperature scaled classifier \cite{guo2017calibration}.

\subsection{MART} 
\citet{Wang2020Improving} observed that addressing more loss on the misclassified sample during training can improve the robustness. Let $f_{\theta}^{(k)}(x)$ as the prediction probability of class $k$ of a given classifier $f_{\theta}$.
For a given sample $(x,y)\sim\mathcal{D}$, the proposed training objective is as follows:
\begin{equation}
    \mathcal{L}_{\mathtt{MART}}:=
    \mathcal{L}_{\mathtt{BCE}}\big(f_{\theta}(x+{\delta}),y\big)
    +    
    \gamma \cdot \big(1-f_{\theta}^{(y)}(x)\big) \cdot \mathtt{KL}\big(f_{\theta}(x) \parallel f_{\theta}(x+{\delta})\big),
    \label{equ:mart}
\end{equation}
where $\mathcal{L}_{\mathtt{BCE}}:=\mathcal{L}_{\mathtt{CE}}(f_{\theta}(x),y)-\log(1-\max_{k\neq y}f_{\theta}^{(k)}(x))$, ${\delta}=\argmax_{\lVert\delta^{'}\rVert_{p}\le\epsilon} \mathcal{L}_{\mathtt{CE}}\big(f_{\theta}(x+\delta^{'}),y\big)$, and $\gamma$ is a hyperparameter. For all experiments, we set the hyperparameter $\gamma=6$ by following the original paper.

\noindent\textbf{MART with consistency regularization.} One can also utilize $\mathcal{L}_{\mathtt{MART}}$ as the base AT objective and adapt our regularization scheme for the final objective $\mathcal{L}_{\mathtt{total}}^{\mathtt{MART}}$. For a given data augmentations $T_{1},T_{2}\sim\mathcal{T}$ and sharpening temperature $\tau$, the final objective is as follows:
\begin{equation}
\begin{split}
\mathcal{L}_{\mathtt{total}}^{\mathtt{MART}} = 
\frac{1}{2} \sum_{i=1}^{2}
\mathcal{L}_{\mathtt{BCE}}\big(f_{\theta}(T_{i}(x)+\delta_{i}),y\big) 
& + 
\gamma \cdot \big(1-f_{\theta}^{(y)}(T_{i}(x))\big) \cdot \mathtt{KL}\big(f_{\theta}(T_{i}(x)) \parallel f_{\theta}(T_{i}(x)+\delta_{i})\big)
\\& + 
\lambda \cdot 
\mathtt{JS} \Big(
\hat{f}_{\theta}\big(T_{1}(x)+\delta_{1};\tau\big) \parallel 
\hat{f}_{\theta}\big(T_{2}(x)+\delta_{2};\tau\big)
\Big)
,
\label{equ:objective-all-mart}
\end{split}
\end{equation}
where $\delta_{i}=\argmax_{\lVert\delta\rVert_{p}\le\epsilon} \mathcal{L}_{\mathtt{CE}}(f_{\theta}\big(T_{i}(x)+\delta),y\big)$, and $\hat{f}_{\theta}(x;\tau)$ is the temperature scaled classifier. 
\section{Detailed Description of Experimental Setups} \label{sec-supple:details}

\textbf {Resource description.} All experiments are processed with a single GPU (NVIDIA RTX 2080 Ti) and 24 instances from virtual CPU (Intel Xeon Silver 4214 CPU @ 2.20GHz).

\noindent\textbf{Dataset description.} For the experiments, we use CIFAR-10, CIFAR-100 \citep{krizhevsky2009learning}, and Tiny-ImageNet.\footnote{The full dataset of CIFAR, and Tiny-ImageNet can be downloaded at \url{https://www.cs.toronto.edu/~kriz/cifar.html} and \url{https://tiny-imagenet.herokuapp.com/}, respectively.} CIFAR-10 and  CIFAR-100 consist of 50,000 training and 10,000 test images with 10 and 100 image classes, respectively. All CIFAR images are 32$\times$32$\times$3 resolution (width, height, and RGB channel, respectively). Tiny-ImageNet contains 100,000 train and 10,000 test images with 200 image classes, and all images are 64$\times$64$\times$3 resolution. For all experiments, we do not assume the existence of a validation dataset.

\section{Algorithm} \label{sec-supple:algorithm}

\begin{algorithm}[ht]
\caption{Consistency Regularization for Adversarial Robustness}
\textbf{Require:} Batch of samples $\mathcal{B}=\{(x_{n},y_{n})\}_{n=1}^{N}$, model $f_{\theta}$, data augmentation family $\mathcal{T}$, classification loss $\mathcal{L}$, regularization hyperparamater $\lambda$, and sharpening temperature $\tau$
\hrule
\begin{algorithmic}[1]
\FORALL{$n \in \{1,...,N\}$}
\STATE Sample $T_{1}, T_{2} \sim \mathcal{T}$ {\color{gray}\# sample two augmentation funtions}
\STATE $(\delta_{1}, \delta_{2})\leftarrow (\argmax_{\lVert\delta\rVert_{p}\le\epsilon} \mathcal{L}(T_{i}(x_{n}),y_{n},\delta;\theta))_{i=1}^{2}$ {\color{gray}\# perturb each augmentation}
\STATE $\mathcal{L}_{\mathtt{adv}}^{(n)}\leftarrow \frac{1}{2}\sum_{i=1}^{2}\mathcal{L}(T_{i}(x_{n}),y_{n},\delta_{i};\theta)$ {\color{gray}\# adversarial training with augmentations}
\STATE $\mathcal{L}_{\mathtt{con}}^{(n)}\leftarrow \mathtt{JS}\Big(
    \hat{f}_{\theta}\big(T_{1}(x_{n})+\delta_{1};\tau\big) \parallel \hat{f}_{\theta}\big(T_{2}(x_{n})+\delta_{2};\tau\big)
    \Big)$ {\color{gray}\# consistency regularization}
\STATE $\mathcal{L}_{\mathtt{total}}^{(n)}\leftarrow \mathcal{L}_{\mathtt{adv}}^{(n)} + \lambda \cdot \mathcal{L}_{\mathtt{con}}^{(n)}$
\ENDFOR
\STATE $\mathcal{L}_{\mathtt{total}} \leftarrow \frac{1}{N}\sum_{n=1}^{N}\mathcal{L}_{\mathtt{total}}^{(n)}$
\STATE $\theta\leftarrow\mathtt{SGD}(\theta,\mathcal{L}_{\mathtt{total}})$
\end{algorithmic}
\end{algorithm}
\setlength{\textfloatsep}{\baselineskip+1pt}

\begin{algorithm}[ht]
\caption{Consistency Regularization with Standard Adversarial Training \cite{madry2018towards}}
\textbf{Require:} Batch of samples $\mathcal{B}=\{(x_{n},y_{n})\}_{n=1}^{N}$, model $f_{\theta}$, data augmentation family $\mathcal{T}$, cross-entropy loss $\mathcal{L}_{\mathtt{CE}}$, regularization hyperparamater $\lambda$, and sharpening temperature $\tau$
\hrule
\begin{algorithmic}[1]
\FORALL{$n \in \{1,...,N\}$}
\STATE Sample $T_{1}, T_{2} \sim \mathcal{T}$ {\color{gray}\# sample two augmentation funtions}
\STATE $(\delta_{1},\delta_{2})
\leftarrow (\argmax_{\lVert\delta\rVert_{p}\le\epsilon} \mathcal{L}_{\mathtt{CE}}(f_{\theta}\big( T_{i}(x_{n})+\delta),y_{n}\big))_{i=1}^{2}$ {\color{gray}\# perturb each augmentation}
\STATE $\mathcal{L}_{\mathtt{adv}}^{(n)}\leftarrow \frac{1}{2}\sum_{i=1}^{2}\mathcal{L}_{\mathtt{CE}}(f_{\theta}\big(T_{i}(x_{n})+\delta_{i}),y_{n}\big)$ {\color{gray}\# standard adversarial training with augmentations}
\STATE $\mathcal{L}_{\mathtt{con}}^{(n)}\leftarrow \mathtt{JS}\Big(
    \hat{f}_{\theta}\big(T_{1}(x_{n})+\delta_{1};\tau\big) \parallel \hat{f}_{\theta}\big(T_{2}(x_{n})+\delta_{2};\tau\big)
    \Big)$ {\color{gray}\# consistency regularization}
\STATE $\mathcal{L}_{\mathtt{total}}^{(n)}\leftarrow \mathcal{L}_{\mathtt{adv}}^{(n)} + \lambda \cdot \mathcal{L}_{\mathtt{con}}^{(n)}$
\ENDFOR
\STATE $\mathcal{L}_{\mathtt{total}} \leftarrow \frac{1}{N}\sum_{n=1}^{N}\mathcal{L}_{\mathtt{total}}^{(n)}$
\STATE $\theta\leftarrow\mathtt{SGD}(\theta,\mathcal{L}_{\mathtt{total}})$
\end{algorithmic}
\end{algorithm}
\setlength{\textfloatsep}{\baselineskip+1pt}

\section{Additional Experimental Results}
\label{sec-supple:more-experiment}

\subsection{Black-box Transfer Attack}
We attempt to test the model under black-box transfer attack, \ie, adversarial examples generated from a different model (typically from a larger model). We test PreAct-ResNet-18 trained under baselines and our regularization loss, with crafted adversarial examples from WideResNet34-10 trained with standard AT (we consider PGD-100 and CW$_{\infty}$ as black-box adversaries). Results in Table \ref{tab:3-main-black-all} demonstrate that our method indeed improves robustness under black-box attacks across baselines. These results not only imply that our regularization method does not suffer from the gradient obfuscation but also show that our method is effective in practical defense scenarios where the target model is hidden from the attackers.
\begin{table}[h]
\caption{
Robust accuracy (\%) of PreAct-ResNet-18 against black-box attacks: adversaries are generated from the standard AT \cite{madry2018towards} pre-trained WideResNet-34-10. All models are trained on CIFAR-10. We use $l_{\infty}$ threat model with $\epsilon=8/255$. The bold indicates the improved results by the proposed method.}
\centering
\small
\begin{tabular}{l cc}
    \toprule
    Method  & PGD-100 & CW$_{\infty}$ \\
    \midrule
    Standard \cite{madry2018towards} &  68.25 & 79.24\\
    \textbf{+ Consistency} & \textbf{74.21} & \textbf{83.19}\\
    \midrule
    TRADES \cite{zhang19p} & 68.96 & 77.60 \\
    \textbf{+ Consistency} & \textbf{70.91} & \textbf{79.77}  \\
    \midrule
    MART \cite{Wang2020Improving} & 69.19 & 77.95 \\
    \textbf{+ Consistency} & \textbf{72.20} & \textbf{79.95} \\
    \bottomrule
\end{tabular}
\label{tab:3-main-black-all}
\end{table}

\subsection{Temperature Scaling.} 
We also investigate the effect of the temperature $\tau$ for the consistency regularization. As shown in Table \ref{tab:6-abla-temp}, the temperature in our method does matter in the robust accuracy of trained models. As our intuition, sharpening the prediction into more sparse, one-hot like distributions with small temperature $\tau<1$ on regularization shows an significant improvement. 
\begin{table}[h]
\caption{Effect of temperature $\tau$ on robust accuracy (\%) against PGD-100, $l_{\infty}$ with $\epsilon=8/255$. We train PreAct-ResNet-18 under CIFAR-10 with consistency regularization loss based on standard AT \cite{madry2018towards}. The bold indicates the best result.}
\small
\centering
\begin{tabular}{l cccccc}
    \toprule
    $\tau$  & 0.5 & 0.8 & 1.0 & 2.0 & 5.0 \\
    \midrule
    PGD-100 & \textbf{56.38} & 56.22 & 55.79 & 56.04 & 55.57 \\
    \bottomrule
\end{tabular}
\label{tab:6-abla-temp}
\end{table}

\subsection{Adversarial Training on $l_{2}$ Constraint Ball}
\label{sec-supple:l2-experiment}

In this subsection, we demonstrate that our regularization scheme is also effective under different types of constraint ball. In particular, we consider adversarial training (AT) on $l_{2}$ constraint ball.

\noindent\textbf{Training details.} We follow the same training configuration as in Section 3, except for the inner maximization constraint and the data augmentation. For inner maximization for all AT, we set the ball radius to $\epsilon=128/255$, step size $\alpha=15/255$, and 10 number of steps with $l_{2}$ constraint. For the augmentation, we simply used color augmentation and Cutout (DeVries and Taylor 2017)over the base augmentation. We use PreAct-ResNet-18 \citep{he2016identity} architecture in all experiments. 

\noindent\textbf{White-box attack.} 
For $l_{2}$ constraint AT, we consider white-box adversarial attacks, including PGD \cite{madry2018towards} with 20 and 100 iterations  (step size with $2\epsilon/k$, where $k$ is the iteration number) and AutoAttack \cite{croce2020reliable}. We report the fully trained model's accuracy and the result of the checkpoint with the best PGD accuracy (of 10 iterations), where each checkpoint is saved per epoch. The results are shown in Table \ref{tab:supple-l2-white-box}. In $l_{2}$ constraint AT, our regularization scheme also significantly improves white-box accuracy against various adversaries and also reduces the overfitting as the best and last robust accuracy gap has been reduced. Interestingly, our regularization scheme also increases the clean accuracy for $l_{2}$ constraint AT. 
This result implies that our method can be a promising method for improving both clean and robust accuracy in various scenarios regardless of the constraint type of the ball.

\noindent\textbf{Unseen adversaries.} We also evaluate our method against \emph{unforeseen} adversaries, \eg, robustness on different attack radii of $\epsilon$ or even on different norm constraints of $l_\infty$ and $l_1$. As shown in Table \ref{tab:supple-l2-unseen}, our regularization technique also consistently, and significantly improves the robustness accuracy against unseen adversaries, in the case of $l_{2}$ constraint AT. We believe our method may also improve the robustness against {unforeseen} attack for other types of constraint ball (\eg, $l_1$). One intriguing direction is to further develop our method toward defensing against multiple perturbations \cite{tramer2019adversarial,maini2020adversarial} which is also an important research field.

\begin{table*}[b]
\caption{
Clean accuracy and robust accuracy (\%) against white-box attacks of networks trained under $l_{2}$ constraint ball. All threat models are $l_{2}$ with $\epsilon=128/255$. Values in parenthesis denote the result of the checkpoint with the best PGD-10 accuracy, where each checkpoint is saved per epoch. The bold indicates the improved results by our regularization loss.}
\small\centering
\begin{tabular}{c l cccc}
    \toprule
    \makecell{Dataset\\(Architecture)} & Method & Clean & PGD-20 & PGD-100 & AutoAttack \\
    \midrule
    \multirow{7}{*}{CIFAR-10}  
    & Standard \cite{madry2018towards} 
    & 90.17 \footnotesize{(89.91)} & 63.61 \footnotesize{(67.93)}
    & 63.36 \footnotesize{(67.77)} & 61.88 \footnotesize{(65.07)} \\
    & \textbf{+ Consistency } 
    & \textbf{91.19} \footnotesize{(87.88)}
    & \textbf{70.03} \footnotesize{(72.77)}
    & \textbf{69.85} \footnotesize{(72.69)}
    & \textbf{68.07} \footnotesize{(70.40)} \\
    \cmidrule{2-6}
    & {TRADES} \cite{zhang19p} 
    & 87.19 \footnotesize{(87.28)} & 65.79 \footnotesize{(70.27)}
    & 65.64 \footnotesize{(70.14)} & 64.28 \footnotesize{(68.14)} \\
    & \textbf{+ Consistency } 
    & \textbf{88.03} \footnotesize{(87.88)}
    & \textbf{72.30} \footnotesize{(72.77)}
    & \textbf{72.23} \footnotesize{(72.69)}
    & \textbf{70.39} \footnotesize{(70.39)} \\
    \cmidrule{2-6}
    & {MART} \cite{Wang2020Improving} 
    & 86.36 \footnotesize{(86.26)} & 64.58 \footnotesize{(68.89)}
    & 64.38 \footnotesize{(68.75)} & 62.63 \footnotesize{(65.66)} \\
    & \textbf{+ Consistency }    
    & \textbf{87.94} \footnotesize{(87.88)}
    & \textbf{71.83} \footnotesize{(72.70)}
    & \textbf{71.73} \footnotesize{(72.53)}
    & \textbf{68.29} \footnotesize{(68.38)} \\
    \midrule
    \multirow{2}{*}{CIFAR-100}
    & Standard \cite{madry2018towards} 
    & 65.94 \footnotesize{(66.26)} & 36.51 \footnotesize{(41.86)} 
    & 36.41 \footnotesize{(41.64)} & 34.98 \footnotesize{(37.79)} \\
    &  \textbf{+ Consistency } 
    & \textbf{67.87} \footnotesize{(66.55)}
    & \textbf{40.00} \footnotesize{(43.33)}
    & \textbf{39.85} \footnotesize{(43.23)}
    & \textbf{37.76} \footnotesize{(39.23)} \\
    \midrule
    \multirow{2}{*}{Tiny-ImageNet} 
    & Standard \cite{madry2018towards} 
    & 55.50 \footnotesize{(56.03)} & 34.49 \footnotesize{(37.19)} 
    & 34.38 \footnotesize{(37.11)} & 33.13 \footnotesize{(34.46)}\\
    & \textbf{+ Consistency }
    & \textbf{56.04} \footnotesize{(58.84)}
    & \textbf{34.95} \footnotesize{(39.54)}
    & \textbf{34.87} \footnotesize{(39.44)}
    & \textbf{33.56} \footnotesize{(36.99)} \\
    \bottomrule
\end{tabular}
\label{tab:supple-l2-white-box}
\end{table*}

\noindent\textbf{Common corruption.} We also validate the effectiveness of our method on corrupted CIFAR-10 dataset \cite{hendrycks2019robustness}. We report the mean corruption error (mCE) of each model in Table \ref{tab:l2-mce}. Our method also shows consistent improvement in corruption dataset even for $l_{2}$ constraint AT. Interestingly,  $l_{2}$ constraint AT shows better performance of mCE compare to $l_{\infty}$ constraint AT across all corruption types. Moreover, the improvement of our method is more significant in $l_{2}$ constraint AT. While $l_{\infty}$ constraint AT's relative improvement is 9.16\%, $l_{2}$ constraint AT shows 14.26\% relative improvement in standard AT \cite{madry2018towards}. 

\begin{table*}[ht]
\caption{Robust accuracy (\%) of PreAct-ResNet-18 trained with $l_{2}$ of $\epsilon=128/255$ constraint against unseen attacks; we use PGD-100 under different sized $l_{2}$ balls and other types of norm balls, \eg, $l_{\infty},$ and $l_{1}$. The bold indicates the improved results by the proposed method.}
\centering\small
\begin{tabular}{c l cc cc cc}
        \toprule
        & & \multicolumn{2}{c}{$l_{2}$} & \multicolumn{2}{c}{$l_{\infty}$} & \multicolumn{2}{c}{$l_{1}$} \\
        \cmidrule(r){3-4} \cmidrule(r){5-6} \cmidrule(r){7-8}
        Dataset & Method$\backslash\epsilon$ & $64/255$ & $256/255$ & $4/255$ & $16/255$ & $2000/255$ & $4000/255$ \\
        \midrule
        \multirow{7}{*}{CIFAR-10}  
        & Standard \cite{madry2018towards}
        & 79.06 & 41.98 & 57.67 & 2.96 & 80.60 & 79.05 \\ 
        & \textbf{+ Consistency }
        & \textbf{82.80} & 41.25 & \textbf{64.76} & \textbf{3.59} & \textbf{82.08} & \textbf{80.92} \\
        \cmidrule{2-8}
        & TRADES \cite{zhang19p} 
        & 78.02 & 41.69 & 61.49 & 10.69 & 78.18 & 77.04 \\
        & \textbf{+ Consistency } 
        & \textbf{81.38} & \textbf{51.22} & \textbf{67.83} & \textbf{12.57} & \textbf{81.36} & \textbf{80.49} \\
        \cmidrule{2-8}
        & MART \cite{Wang2020Improving} 
        & 77.27 & 42.30 & 59.22 & 5.11 & 77.64 & 76.44 \\
        & \textbf{+ Consistency }
        & \textbf{80.91} & \textbf{49.53} & \textbf{67.61} & \textbf{6.38} & \textbf{80.71} & \textbf{79.45}\\
        \midrule
        \multirow{2}{*}{CIFAR-100}  
        & Standard \cite{madry2018towards}
        & 51.34 & 16.65 & 31.00 & 1.61 & 54.32 & 52.82 \\
        & \textbf{+ Consistency }
        & \textbf{53.83} & \textbf{19.64} & \textbf{34.54} & \textbf{1.83} & \textbf{54.33} & \textbf{53.00} \\
        \midrule
        \multirow{2}{*}{Tiny-ImageNet}  
        & Standard \cite{madry2018towards}
        & 44.95 & 18.65 & 14.27 & 0.22 & 51.35 & 51.04 \\
        & \textbf{+ Consistency }
        & \textbf{45.63} & \textbf{19.63} & \textbf{14.93} & \textbf{0.29} & \textbf{51.69} & \textbf{51.06} \\
        \bottomrule
\end{tabular}
\label{tab:supple-l2-unseen}
\end{table*}

\begin{table}[t]
\centering\small
\caption{Mean corruption error (mCE) (\%) of PreAct-ResNet-18 trained on CIFAR-10 under $l_{2}$ constraint ball, and tested with CIFAR-10-C dataset \cite{hendrycks2019robustness}. The bold indicates the improved results by the proposed method.}
\centering
\small
\begin{tabular}{l c}
    \toprule
    Method  & mCE $\downarrow$  \\
    \midrule
    Standard \cite{madry2018towards} &  17.81 \\
    \textbf{+ Consistency} & \textbf{15.27} \\
    \midrule
    TRADES \cite{zhang19p} & 20.55 \\
    \textbf{+ Consistency} & \textbf{18.21} \\
    \midrule
    MART \cite{Wang2020Improving} & 21.42  \\
    \textbf{+ Consistency} & \textbf{18.38} \\
    \bottomrule
\end{tabular}
\label{tab:l2-mce}
\end{table}

\subsection{Variance of Results Over Multiple Runs}

In Table \ref{tab:supple-seed}, we report the mean and standard deviation of clean and robust accuracy of CIFAR-10 results for standard AT \cite{madry2018towards}, and our method. We observe both accuracies of a given training method are fairly robust to network initialization. 

\begin{table}[h]
\caption{
Clean accuracy and robust accuracy (\%) against white-box attacks of networks trained under CIFAR-10. All threat models are $l_{\infty}$ with $\epsilon=8/255$. The reported values are the mean and standard deviation across 5 seeds. The bold indicates the improved results by our regularization loss.}
\small\centering
\begin{tabular}{l ccccc}
    \toprule
    Method & Clean & PGD-20 & PGD-100 & CW$_{\infty}$ & AutoAttack \\
    \midrule
    Standard \cite{madry2018towards} 
    & 84.44\stdv{0.34} & 
    45.85\stdv{0.25} & 
    45.67\stdv{0.27} & 
    44.91\stdv{0.28} & 
    40.71\stdv{0.28} \\
    \textbf{+ Consistency} 
    & \textbf{86.24}\stdv{0.23} 
    & \textbf{56.50}\stdv{0.27} 
    & \textbf{56.24}\stdv{0.22} 
    & \textbf{52.32}\stdv{0.22} 
    & \textbf{48.87}\stdv{0.14} \\
    \bottomrule
\end{tabular}
\label{tab:supple-seed}
\end{table}

\section{Additional Analysis on the Consistency Regularization Loss}

\subsection{Design Choices of Discrepancy Function in the Consistency Regularization Loss}
We examine two other popular designs of discrepancy function for the consistency regularization instead of Jensen-Shannon divergence, namely, mean-squared-error and KL-divergence as follow:
\begin{equation}
\mathcal{L}_{\mathtt{MSE}}:=\Big\lVert f_{\theta}\big(T_{1}(x)+\delta_{1}\big) - f_{\theta}\big(T_{2}(x)+\delta_{2}\big) \Big\rVert^{2}_{2},
\label{eq:mse}
\end{equation}
\begin{equation}
\mathcal{L}_{\mathtt{KL}}:= \mathtt{KL}\Big(f_{\theta}\big(T_{1}(x)+\delta_{1}\big) \parallel f_{\theta}\big(T_{2}(x)+\delta_{2}\big)\Big),
\label{eq:kl}
\end{equation}
where $\delta_{i}$ is the adversarial noise of $T_{i}(x)$. We use the same setup as in Section 3 and jointly train the TRADES \cite{zhang19p} objective with different choices of consistency losses. The results are presented in Table \ref{tab:loss}. In general, we observe that the $\mathcal{L}_{\mathtt{MSE}}$ regularizer can improve both clean and robust accuracies, but it could not achieve better robustness than the Jensen-Shannon divergence. Moreover, we observed that $\mathcal{L}_{\mathtt{KL}}$ significantly degrade the robustness against the AutoAttack \cite{croce2020reliable}. The interesting point is that the  $\mathcal{L}_{\mathtt{KL}}$ regularizer extremely lowers the confidence (\ie, maximum softmax probability) of the classifier, \eg, average confidence of the clean examples is 0.37 while other model's confidences are larger than 0.56. Based on our empirical finding, we conjecture that models with very low confidence lead to the venerability against the AutoAttack.

\begin{table}[h]
\caption{Comparison with different of discrepancy functions under TRADES \cite{zhang19p}. Clean accuracy and robust accuracy (\%) against white-box attacks of PreAct-ResNet-18 trained on CIFAR-10. We use $l_{\infty}$ threat model with $\epsilon=8/255$. The bold indicates the best result.}
\centering\small
\vspace{0.05in}
\begin{tabular}{l ccc}
\toprule
Discrepancy & Clean & PGD-100 & AutoAttack \\
\midrule
None \eqref{equ:trades} & 82.87 & 50.83 & 46.32 \\
MSE \eqref{eq:mse} & 83.05 & 54.32 & 47.30 \\
KL-div \eqref{eq:kl} & 82.70 & 53.58 & 43.25 \\
JS-div \eqref{equ:objective-all-trades} & \textbf{83.63} & \textbf{55.00} & \textbf{47.68} \\
\bottomrule
\end{tabular}
\label{tab:loss}
\end{table}

\subsection{Comparison with \citet{hendrycks2019augmix}}
Our consistency regularization can be seemingly similar to AugMix \cite{hendrycks2019augmix}, however, as mentioned in Section 2, there exists a fundamental difference; our method match the prediction after attacking the `clean` augmented samples independently. We also find that there exist other differences between AugMix which plays an important role in preventing overfitting. To be specific, when utilizing base augmentation (\ie, random cropping and flipping) in the regularization loss, it may induce robust overfitting even we attack the `clean` augmented instances. 

To this end, we extend AugMix loss to match the attack direction of the given instance. Concretely, for a given sample $(x,y)$, $T\sim\mathcal{T}_{\mathtt{base}}$ and $T_{1},T_{2}\sim\mathcal{T}_{\mathtt{AutoAug}}$\footnote{AugMix also  propose a new augmentation scheme (\ie, mixing augmentation), nonetheless, we only focus on the consistency loss (we observed that mixing augmentation shows similar performance to AutoAugment in AT)}, the extension version of AugMix is as follows: 
\begin{equation}
\mathtt{JS}\Big(f_{\theta}\big(T(x)+\delta\big) \parallel f_{\theta}\big(T_{1}(x) +\delta_{1}\big) \parallel f_{\theta}\big(T_{2}(x) +\delta_{2}\big) \Big),
\label{equ:augmix} 
\end{equation}
where $\mathtt{JS}$ indicates the Jensen-Shannon divergence, $\delta,\delta_{1},\delta_{2}$ is the adversarial noise of $T(x),T_{1}(x),T_{2}(x)$, respectively. As shown in Table \ref{tab:da-overfit}, we find utilizing base augmentation in to the consistency loss, only shows marginal improvement on the adversarial robustness compared to ours. We conjecture that not exposing base augmentations, is crucial for designing the consistency regularization scheme in AT.
\begin{table}[h]
\centering\small
\caption{Comparison under different training epoch of standard AT \cite{madry2018towards} with our consistency loss. Last and best robust accuracy (\%), against PGD-100 of PreAct-ResNet-18 trained on CIFAR-10. We use $l_{\infty}$ threat model with $\epsilon=8/255$.}
\begin{tabular}{lcc}
\toprule
Method & Best & Last \\
\midrule
Standard \cite{madry2018towards} & 52.67 & 44.86 \\ 
+ AugMix \eqref{equ:augmix} \cite{hendrycks2019augmix} & 52.87 & 48.08 \\  
\textbf{+ Consistency } & \textbf{57.39} & \textbf{56.38} \\
\bottomrule
\end{tabular}
\label{tab:da-overfit}
\end{table}

\subsection{Runtime Analysis}
One might concern the training cost of our method, as it is being doubled compared to baseline AT methods due to the two independent adversarial examples. However, we found that our method maintains almost the same robust accuracy even under the same computational budget as the baselines by reducing the training epochs in half. To this end, we train standard AT \cite{madry2018towards} objective jointly with our regularization loss under CIFAR-10. As shown in Table \ref{tab:runtime}, the gap of robust accuracy (between 100 and 200 epoch trained models) under PGD-100, and AutoAttack is only 0.28\% and 0.07\%, respectively.

\begin{table}[h]
\caption{Comparison under different training epoch of standard AT \cite{madry2018towards} with our consistency loss. Robust accuracy (\%), against white-box attacks of PreAct-ResNet-18 trained on CIFAR-10. We use $l_{\infty}$ threat model with $\epsilon=8/255$.}
\centering\small
\begin{tabular}{lcc}
\toprule
Epoch & PGD-100 & AutoAttack \\
\midrule
100 & 56.10 & 48.50 \\
200 & 56.38 & 48.57 \\
\bottomrule
\end{tabular}
\label{tab:runtime}
\end{table}

\section{Additional Analysis on the Data Augmentations}

\subsection{Ablation Study on Data Augmentations for Adversarial Training}
\begin{figure}[ht]
\centering
\includegraphics[width=0.45\textwidth]{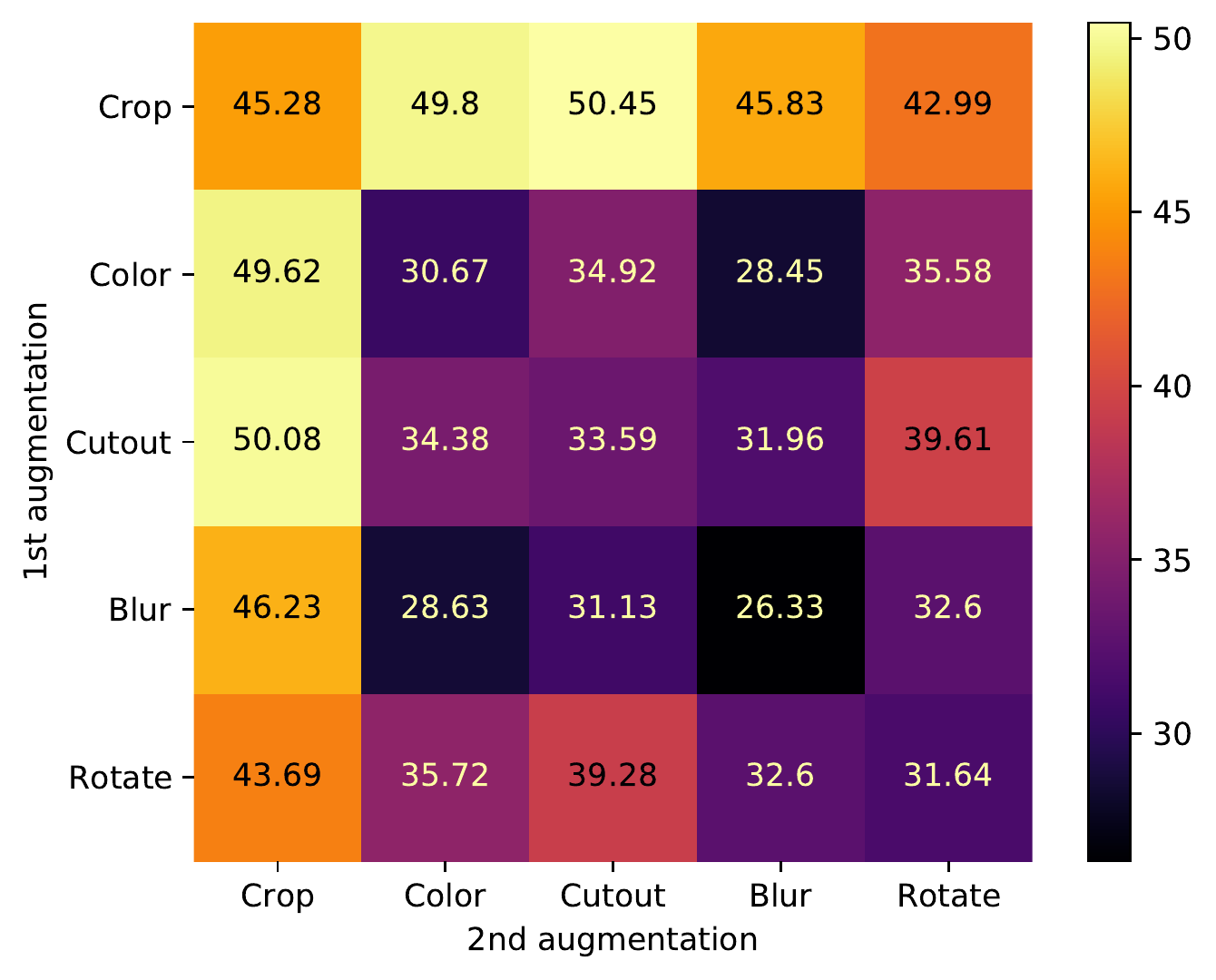}
\caption{Robust accuracy (\%) against PGD-10 attack of PreAct-ResNet-18 trained with standard AT \cite{madry2018towards}. We use $l_{\infty}$ threat model with $\epsilon=8/255$. We train on CIFAR-10 dataset under individual or composition of augmentations. The diagonal implies the single augmentation, and the off-diagonal indicates the results of two augmentations being performed in sequence.
}\label{fig:augment-composition}
\end{figure}

\noindent We attempt to discover which augmentations family improves the generalization in adversarial training (AT). To this end, we select a composition of augmentations that are positively correlated with each other: we combine the augmentations in pairs and apply for AT. We use the same experimental setup from Section \ref{sec:observation} and report the robust accuracy of the final model. We consider various types of augmentations, including random crop (with horizontal flip), rotation \cite{gidaris2018unsupervised} (consist of 90\degree, 180\degree, and 270\degree), Cutout \cite{devries2017improved}, color distortion (color jitter and grayscale), and Gaussian blur. As shown in Figure \ref{fig:augment-composition}, combining random crop (with horizontal flip) with other augmentations such as color distortion and Cutout further improve the robustness. In particular, Color-jitter and Cutout were crucial, which diversely transform the sample than other DAs; we hypothesize that such sample diversity through augmentations is significant for improving the robustness.

\subsection{Learning Dynamics of Adversarial Training with Additional Augmentations}
\begin{figure*}[ht]
\centering
\begin{subfigure}{0.42\textwidth}
\centering
\includegraphics[width=\textwidth]{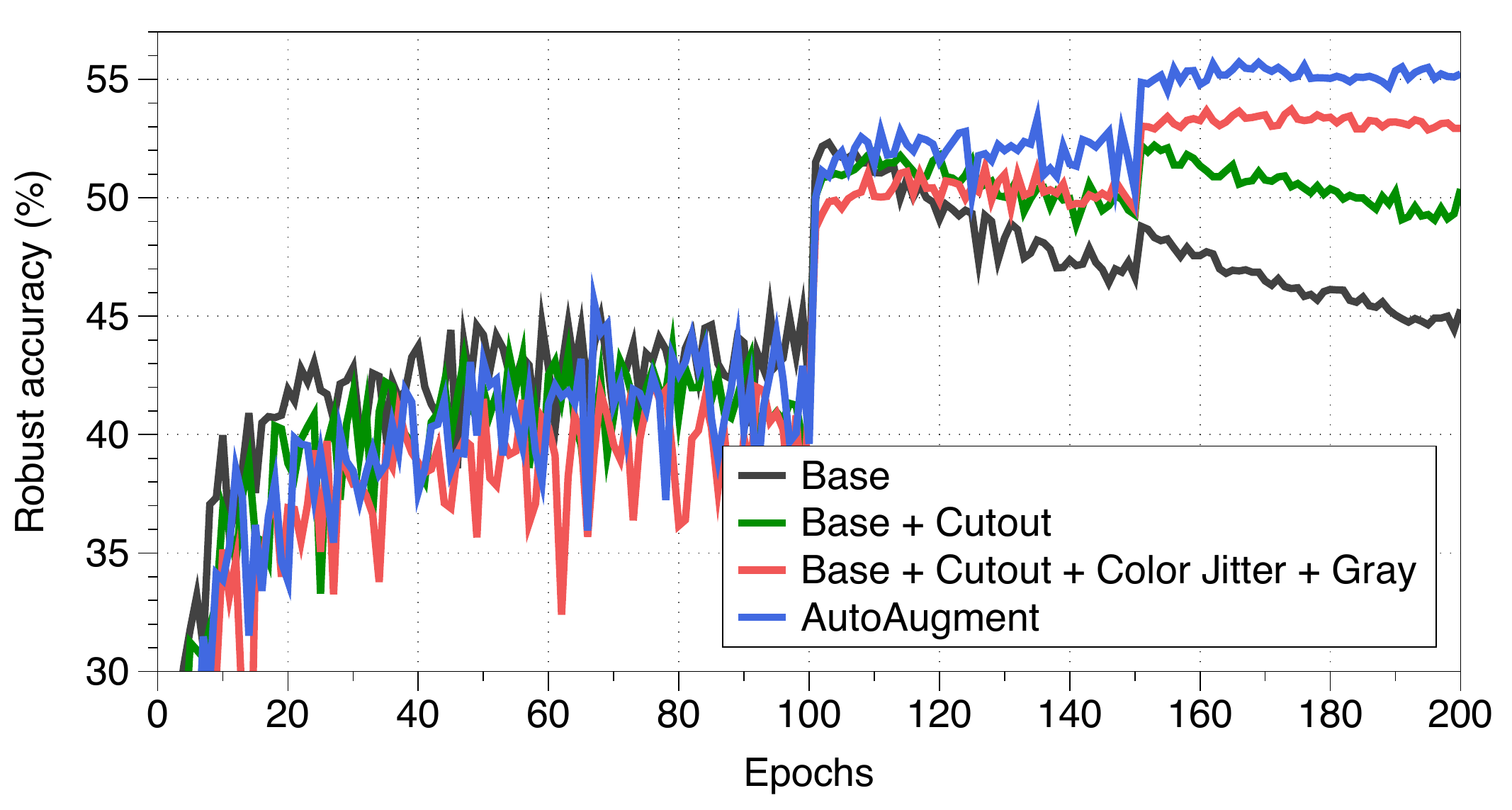}
\caption{Robust accuracy}
\label{fig:supple-augmentation-robust}
\end{subfigure}
~~~
\begin{subfigure}{0.42\textwidth}
\centering
\includegraphics[width=\textwidth]{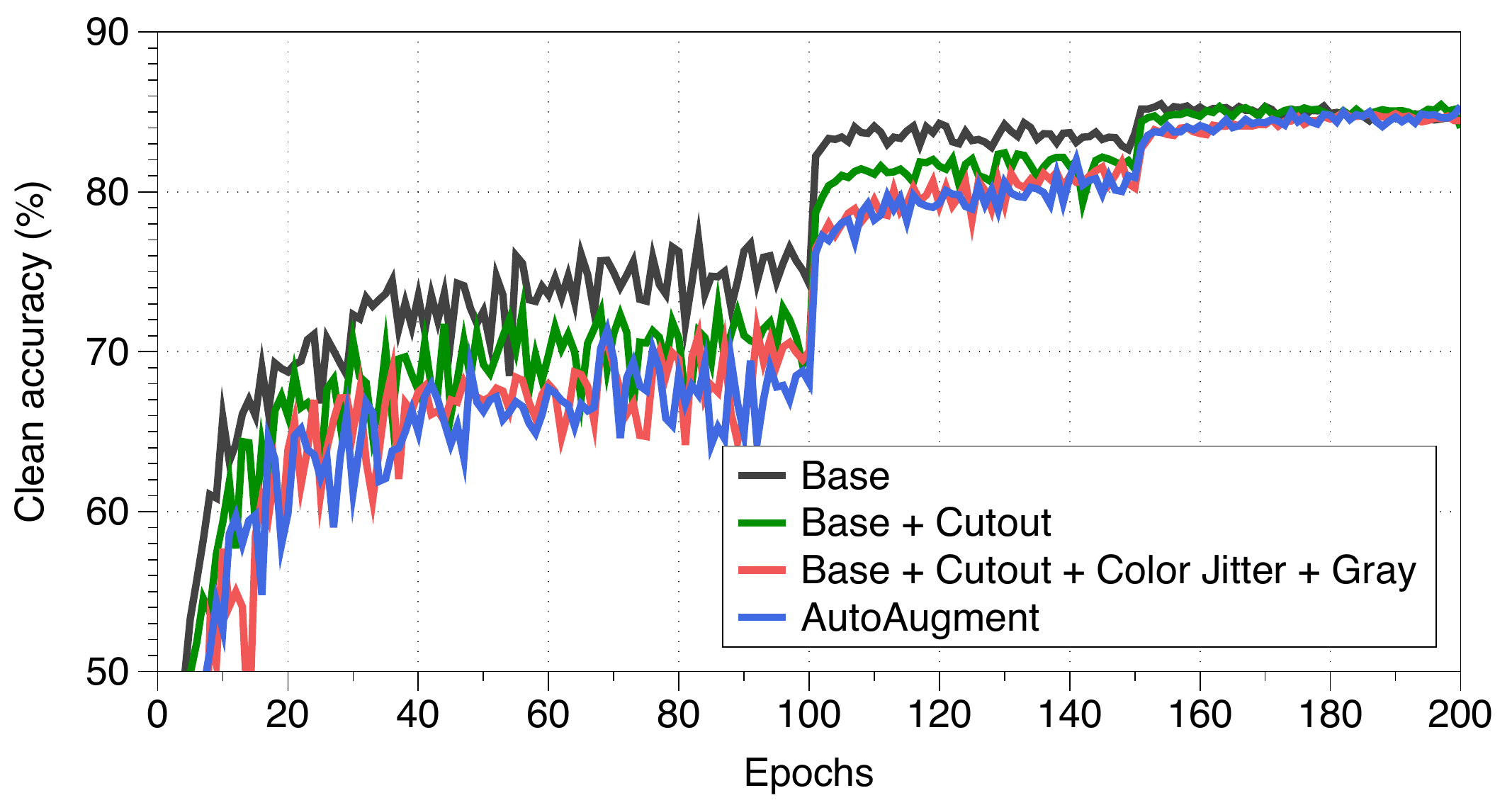}
\caption{Clean accuracy}
\label{fig:supple-augmentation-clean}
\end{subfigure}
\caption{
Clean accuracy and robust accuracy (\%) against PGD-10 attack of standard AT \cite{madry2018towards} on different augmentations with PreAct-ResNet-18 under CIFAR-10. We use $l_{\infty}$ threat model with $\epsilon=8/255$. Base indicates the random crop and horizontal flip.
}
\label{fig:supple-augmentation}
\end{figure*}

\noindent Figure \ref{fig:supple-augmentation} shows the test robust and clean accuracy of standard AT \cite{madry2018towards} with additional augmentations from the common practice (\ie, random crop and horizontal flip). We denote such common practice augmentation set as base augmentation. As shown in Figure \ref{fig:supple-augmentation-robust}, further use of Cutout \cite{devries2017improved} from the base augmentation improves robust accuracy. However, one can observe that it still overfits in the end. In contrast, additional color augmentations can train the classifier without overfitting. Further utilizing more diverse augmentations, \ie, AutoAugment, the overfitting issue significantly reduces and even the best robust accuracy improves. We observed that the augmentation choice does not affect the clean accuracy much. Nonetheless, by optimizing our regularization loss simultaneously, the clean accuracy significantly improves (see Table \ref{tab:supple-l2-white-box}). 

\subsection{Discussion with Recent Works in Data Augmentations for Adversarial Training}
In this work, we observed that the data augmentation (DA) effectively prevents the robust overfitting which is contrast to the recent studies \cite{gowal2020uncovering,rebuffi2021fixing,rebuffi2021data} which reported AutoAugment does not help for AT. We recognize the authors utilize the TRADES \cite{zhang19p} objective as a basic formula for the observation while we basically used standard AT \cite{madry2018towards} objective. We find TRADES under strong augmentations can significantly lower the classifier's confidence (\ie, maximum softmax probability) and shows low robust accuracy, \eg, 46.32\% with base augmentation and 46.03\% with AutoAugment on PreAct-ResNet-18 under AutoAttack. We conjecture that AutoAugment act as a hard example for TRADES, hence, lead the classifier to predict low confidence. Based on this observation, we find that scheduling the augmentation policy for TRADES improves the robust accuracy, \eg, 48.50\%. Specifically, we applied base augmentation until 100 epochs then trained with AutoAugment for the rest of the training.

\end{document}